\documentclass[nohyperref]{article}

\usepackage{microtype}
\usepackage{graphicx}
\usepackage{subcaption}
\usepackage{booktabs} 

\usepackage{hyperref}

\usepackage[accepted]{icml2022}

\usepackage{amsmath}
\usepackage{amssymb}
\usepackage{mathtools}
\usepackage{amsthm}
\usepackage{makecell}

\usepackage[capitalize,noabbrev]{cleveref}

\theoremstyle{plain}

\theoremstyle{definition}

\theoremstyle{remark}

\usepackage[textsize=tiny]{todonotes}

\usepackage{pifont}
\newcommand{\cmark}{\ding{51}}%
\newcommand{\xmark}{\ding{55}}%

\usepackage[splitrule]{footmisc}

\icmltitlerunning{Persistently Trained, Diffusion-assisted Energy-based Models}

\begin{document}

\twocolumn[
\icmltitle{Persistently Trained, Diffusion-assisted Energy-based Models}



\icmlsetsymbol{equal}{*}

\begin{icmlauthorlist}
\icmlauthor{Xinwei Zhang}{rutgers}
\icmlauthor{Zhiqiang Tan}{rutgers}
\icmlauthor{Zhijian Ou}{tsinghua}
\end{icmlauthorlist}

\icmlaffiliation{rutgers}{Department of Statistics, University of Rutgers, Piscataway, US}
\icmlaffiliation{tsinghua}{Department of Electronic Engineering, Tsinghua University, Beijing, China}

\icmlcorrespondingauthor{Zhiqiang Tan}{ztan@stat.rutgers.edu}
\icmlcorrespondingauthor{Zhijian Ou}{ozj@tsinghua.edu.cn}

\icmlkeywords{Machine Learning, ICML}

\vskip 0.3in
]



\printAffiliationsAndNotice{}  

\newcommand{\fpartial}[2]{\frac{\partial #1}{\partial #2}}

\def\dif{\mathrm d}
\def\E{\mathbb{E}}

\def\me{\mathrm e}
\def\N{\mbox{N}}

\def\T{ {\mathrm{\scriptscriptstyle T}} }

\begin{abstract}
Maximum likelihood (ML) learning for energy-based models (EBMs)
is challenging, partly due to non-convergence of Markov chain Monte Carlo.
Several variations of ML learning have been proposed, but existing methods all fail to achieve
both post-training image generation and proper density estimation.
We propose to introduce diffusion data and learn a joint EBM, called diffusion assisted-EBMs, through persistent training
(i.e., using persistent contrastive divergence) with an enhanced sampling algorithm to properly sample from complex, multimodal distributions.
We present results from a 2D illustrative experiment and image experiments and demonstrate that,
for the first time for image data, persistently trained EBMs can {\it simultaneously} achieve
long-run stability, post-training image generation, and superior out-of-distribution detection.
\end{abstract}

\section{Introduction} \label{sec:introduction}

Energy-based models (EBMs) parameterize an unnormalized data density or
equivalently an energy function, defined as the negative log density up to an additive constant.
There has been persistent and ongoing interest in developing effective modeling and training techniques for learning EBMs from
complex data such as natural images.
A partial list of examples include
\citet{lecun2006tutorial}, \citet{xie2016theory}, \citet{du2019implicit}, \citet{nijkamp2019shortrun}, and \citet{grathwohl2020JEM}.
Apart from maximum likelihood (ML) learning, various other training principles are available,
including score matching \citep{hyvrinen2005scorematch, saremi2018deepenergy}, noise-contrastive estimation \citep{gutmann2012ncejmlr},
and $f$-divergence minimization \citep{yu2020fdiverg}.
See Table 1 in \citet{grathwohl2021nomcmc} for a comparison.

\begin{table}[t] 
\caption{Comparison of training methods for EBMs}
\label{tb:property_summary}
\vskip .05in
\begin{minipage}{\columnwidth}
\begin{center}
\begin{small}
\begin{sc}
\resizebox{\columnwidth}{!}{
\begin{tabular}{ c | c c c   }\toprule
 &  \makecell{Long-run}& \makecell{Post} & Glob. Energy \\\midrule
EBM Pers. & \cmark & \xmark & \xmark\\
EBM CD    & \xmark & \xmark & \xmark\\
EBM Noise  & \xmark & \cmark & \xmark\\
EBM Hybrid & \xmark & \cmark & \xmark\\
Dif. Rec. Lik. ({\normalfont\scriptsize DRL}) & \cmark/\xmark\footnote{Mixed results are obtained in our experiments.} & \cmark & \xmark\\
DA-EBM ({\normalfont ours})& \cmark & \cmark & \cmark\\ \bottomrule
\end{tabular}%
}
\end{sc}
\end{small}
\end{center}
\end{minipage}
\vskip -0.1in
\end{table}

We aim to advance ML learning for EBMs.
Several variations of ML learning have been proposed,
and impressive performances have been reported \cite{du2019implicit,gao2020Diffusion}.
See Section~\ref{sec:background} for a discussion of initialization schemes
for Markov chain Monte Carlo (MCMC) during training, including persistent, data, noise, and hybrid initializations.
However, training EBMs remains challenging and complicated by conflicting ideas.
As shown in Table \ref{tb:property_summary},
existing training methods all fail to achieve at least one of the desired learning properties, defined as follows.\vspace{-.15in}
\begin{itemize}\addtolength{\itemsep}{-.1in}
\item Long-run stability (LONG-RUN): Long-run MCMC samples generated (after training) using the learned energy starting from {\it real images} remain realistic.

\item Post-training sampling (POST): MCMC samples generated (after training) using the learned energy starting from {\it random noises} are realistic.

\item Global energy estimation: The learned energy function is globally aligned between different modes separated by low-density regions.
A learned energy may lead to long-run stability, but {\it only} be locally meaningful (i.e., accurate near local modes). See Figure \ref{fig:1d_example}.

\end{itemize}\vspace{-.15in}
For the noise initialization method, the learned short-run MCMC can be used to generate realistic images from random noises
similarly in GAN \cite{goodfellow2014gan} or Glow \cite{kingma2018glow}
but the learned energy functions seem to be invalid due to the lack of long-run stability \cite{nijkamp2020anatomy}.
Persistent training can be implemented using certain training strategies
to produce stable long-run samples from real data \cite{nijkamp2020anatomy}, but this approach
has not been successful in generating new realistic images  from random noises after training.

Current applications of EBMs to out-of-distribution (OOD) detection \cite{du2019implicit, grathwohl2020JEM}
leave much room for improvement due to the OOD reversal phenomenon, where higher likelihoods are assigned to
OOD observations than in-distribution (InD) observations.
Such phenomena are observed in both flow-based deep generative modeling \citep{nalisnick2019dgmood, choi2019waic}
and usual and joint EBMs \cite{grathwohl2020JEM}.
A possible explanation and remedy were proposed in \citet{nalisnick2019oodtypicality}, using the concept of typicality \cite{Cover2006ElementsofIT}.
An alternative hypothesis underlying our work is that
the OOD reversal may occur because the estimated energies (or log-likelihoods) are locally meaningful, but not globally aligned,
so that higher likelihoods may be assigned to OOD data points near one mode than InD data points near another mode.
See the 1D example in Section~\ref{sec:diagnosis}.

In this work, we make three main contributions. First, we identify and explain, for the first time, the phenomenon that
for complex, multimodal data distributions, persistent training using an MCMC sampler which suffers local mixing may
only learn local energy functions, which are locally meaningful but globally misaligned.
Second, motivated by this understanding and inspired by recent success of
score-based diffusion modeling,
we propose to introduce diffusion data through a forward diffusion process and learn a joint EBM, called diffusion assisted-EBMs (DA-EBMs),
from both the original training data and diffusion data at different time steps. We pursue persistent training while incorporating
an enhanced sampling algorithm to overcome local mixing \cite{tan2017mixturesampling}.
See Section~\ref{sec:comparison} for a comparison of our method with score-based diffusion modeling \citep{sohl2015deep, song2019denoising, ho2020ddpm, song2021sde}
and diffusion recovery likelihood \cite{gao2020Diffusion}.
Third, we present results from a 2D illustrative experiment and image experiments on MNIST-type data and demonstrate that,
for the first time for image data, persistently trained EBMs can {\it simultaneously} achieve
long-run stability, post-training image generation, and superior OOD detection,
indicating that the learned energy is globally more meaningful than previously obtained.  Our code is available at {\footnotesize \url{https://github.com/XinweiZhang/DAEBM}.}

\section{Background: Energy-based Models} \label{sec:background}


An energy-based model (EBM) is defined in the form \vspace{-.05in}
\begin{align} \label{eq:ebm}
p_\theta(x) = \frac{\exp(-U_\theta(x))}{Z(\theta)}
\end{align}
where $U_\theta(x)$ is an energy (or potential) function, for example, specified as a neural network with parameter $\theta$ and $Z(\theta) = \int \exp( - U_{\theta}(x)) \dif x$ is the normalizing constant. Traditionally, EBMs are used for unsupervised image modeling, where $x$ denotes an image configuration \citep{lecun2006tutorial, xie2016theory, du2019implicit, nijkamp2019shortrun}. Recently, EBMs are also considered in supervised or semi-supervised settings, where $x$
is a pair of an image configuration and a class label \cite{ou2018INF, grathwohl2020JEM}.
Such models are called joint EBMs.

Statistically, EBMs can be trained by maximum likelihood (ML). Given training data $\mathcal{D}=\{z_i\}_{i=1}^n$, the gradient of the log-likelihood $ l(\theta) = \frac{1}{n}\sum_{i=1}^n \log p_\theta(z_i)$ is\vspace{-.1in}
\begin{align}\label{eq:ebm-grad}
   \fpartial{}{\theta} l(\theta)  = -  \E_{\hat p} \Big\{\fpartial{}{\theta} U_\theta(x) \Big\}
   + \E_{p_\theta} \Big\{ \fpartial{}{\theta} U_\theta(x) \Big\},
\end{align}
where $\hat p$ denotes the empirical distribution on $\mathcal{D}$ and
$\E_q (\cdot)$ denotes the expectation with respect to $q$.
The ML estimator of $\theta$ 
is obtained as a solution to $\fpartial{}{\theta}  l(\theta) =0$. 
However, a challenge is that the expectation $\E_{p_\theta} \{ \fpartial{}{\theta} U_\theta(x) \}$ in (\ref{eq:ebm-grad}) is usually difficult to evaluate.
This issue can be addressed through stochastic approximation (SA), which provides
a principled and flexible approach for numerically solving intractable equations \cite{robbins1951sa, benveniste1990sa}.
Given current value $\theta_0$ and replay buffer $\mathcal{B}$ (which stores current synthetic data),
an SA iteration for ML estimation (ML-SA) performs the following operations. \vspace{-.1in}
\begin{itemize}\addtolength{\itemsep}{-.1in}
    \item Sampling: Draw $x_1$ from $\mathcal{D}$, $\tilde x_0$ from $\mathcal{B}$, and $\tilde x_1$ from a Markov transition kernel $K_{\theta_0}(\tilde x_0, \cdot)$,
    where $K_{\theta}(\cdot,\cdot)$ is assumed to leave $p_{\theta}$ invariant (among other technical conditions);
    reset $\tilde x_0$ to $\tilde x_1$ in $\mathcal{B}$.

    \item Updating: Update $\theta$ by gradient ascent as
\begin{align} \label{eq:ebm-update}
\hspace*{-.1in} \theta_1 = \theta_0 + \gamma \Big\{ - \fpartial{}{\theta} U_\theta(x_1) + \fpartial{}{\theta} U_\theta(\tilde{x}_1) \Big\} \Big|_{\theta=\theta_0},
\end{align}
where $\gamma$ is a learning rate.
\end{itemize} \vspace{-.15in}
The two operations are also called synthesis and analysis \cite{xie2016theory}.
The Markov transition from $\tilde x_0$ to $\tilde x_1$ can be defined by multiple (more elementary) sampling steps such as (\ref{eq:sgld}) below.
Moreover, multiple observations can be allowed in each iteration by drawing a mini-batch of real observations from $\mathcal D$ and
running multiple parallel chains to draw new synthetic observations and returning them to $\mathcal B$.

There are two important aspects of the sampling operation, where different choices may lead to variations related to but distinct from the ML-SA algorithm.
One is the choice of the Markov transition $K_{\theta}(\cdot,\cdot)$, represented by a Markov chain Monte Carlo (MCMC) algorithm.
A popular MCMC algorithm is
Langevin sampling. Given current observation $\tilde x_0$, a new observation is proposed as \vspace{-.05in}
\begin{align}\label{eq:sgld}
    \tilde x_{1/2} = \tilde  x_0 - (\sigma^2/2) \nabla_x U_{\theta_0} (\tilde x_0) + \sigma \varepsilon,
\end{align}
where $\varepsilon \sim \N(0, I)$ is a Gaussian noise, and $\sigma$ is a step size.
The next observation $\tilde x_1$ may directly be the proposal $\tilde x_{1/2}$,
in which case the Markov transition from $\tilde x_0$ to $\tilde x_1$ does not strictly leave $p_{\theta_0}$ invariant,
but the sampling bias may usually be small for $\sigma \approx 0$.
To allow large $\sigma$, a correction can be achieved
by accepting or rejecting the proposal $\tilde x_{1/2}$, i.e., setting $\tilde x_1= \tilde x_{1/2}$ or $\tilde x_0$,
with the Metropolis--Hastings probability.
Langevin sampling with rejection is known as the Metropolis-Adjusted Langevin Algorithm (MALA) \citep{besag1994mala, roberts1996exponential}.

The other choice in the sampling operation is the initialization scheme, i.e.,
the choice $\tilde x_0$ used in $K_{\theta_0}(\tilde x_0, \cdot)$ above.
We summarize existing initialization schemes below.
Note that the initialization here refers to how the starting value is chosen for the Markov transition during training,
not how the starting value is chosen  in long-run MCMC sampling after training or in  post-training image generation. \vspace{-.15in}
\begin{itemize}\addtolength{\itemsep}{-.1in}
    \item Persistent initialization: Markov chains are started from past synthetic observations in the previous training iteration.
    This is the rule prescribed in the ML-SA algorithm and also known as Persistent contrastive divergence (PCD) \citep{tieleman2008}.
    \citet{nijkamp2020anatomy} discussed various tuning choices such as the Langevin step size $\sigma$ and learning rate $\gamma$ to
    obtain stable long-run samples for persistent training.

    \item Data initialization: \citet{hinton2002rbm} proposed contrastive divergence (CD), where Markov chains are initialized by real training data and run for several steps to obtain synthetic data. \citet{gao2020Diffusion} used Markov chains initialized by noise-added real data to train EBMs through recovery likelihood.
    \item Noise initialization: \citet{nijkamp2019shortrun,nijkamp2020anatomy} studied training EBMs with short-run MCMC, where Markov chains are always started from
    random noises and run for a fixed number of Langevin steps (\ref{eq:sgld}). 
    \item Hybrid initialization: Several works employed a hybrid of persistent initialization and noise initialization, that is, initializing Markov chains either by past synthetic observations at a certain rate (e.g., 95\%) or, in the remaining time, by random noises \citep{du2019implicit, grathwohl2020JEM}.
    \item Ancillary generator: Because MCMC sampling may be computationally costly and inefficient, several works proposed
    training an ancillary generator to initialize Markov chains to reduce MCMC steps needed or even directly generate synthetic samples to avoid MCMC \citep{kim2016deep, xie2018cooperative, ou2018INF, dai2019exponential, grathwohl2021nomcmc}.
\end{itemize}\vspace{-.15in}

We point out that from our experiment results (Sections~\ref{sec:toy_example}--\ref{sec:mnist}),
hybrid initialization (although previously treated as a close variation of PCD) behaves similarly to noise initialization rather than to persistent initialization.

\section{Proposed method}  \label{sec:proposed}

In spite of recent progress, learning EBMs for complex data like natural images remains challenging with various dilemmas (see Table \ref{tb:property_summary}).
There are two aims in our investigation: (i) to further study the behavior of persistent training and
(ii) to develop a new method by leveraging diffusion data for persistent training to achieve satisfactory learning outcomes
in several aspects {\it simultaneously}, including long-run stability, post-training image generation, and OOD detection.

\begin{figure}[!t]
\begin{center}
\centerline{\includegraphics[width=\columnwidth]{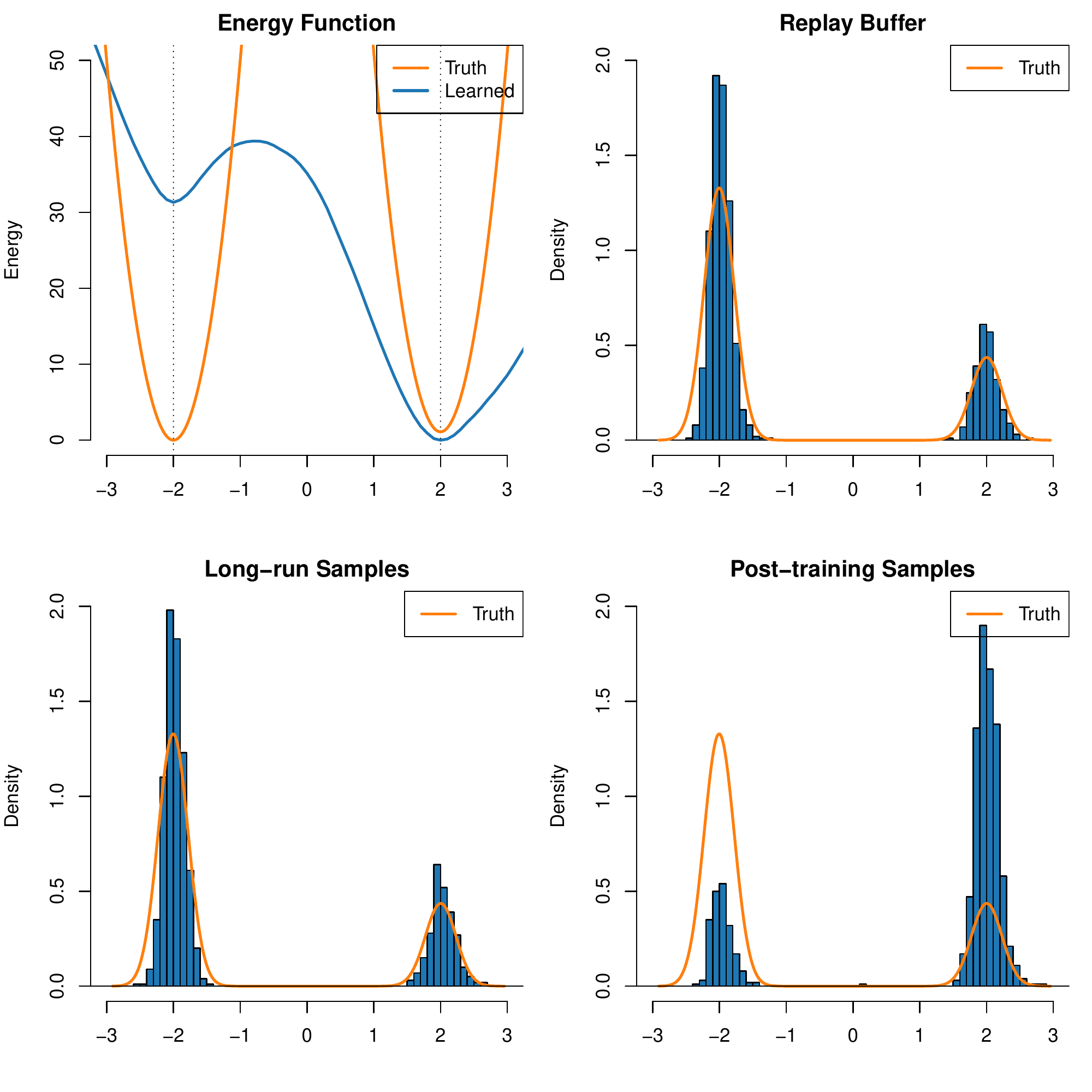}} \vspace{-.2in}
\caption{Results from 1D example}
\label{fig:1d_example}
\end{center}
\vskip -0.4in
\end{figure}

\subsection{Diagnosis: local mixing, local energy} \label{sec:diagnosis}

We present a simple but informative 1D example.
The training data consist of 750 observations from $\N(-2, .1^2)$ and 250 observations from $\N(2, .1^2)$. The EBM
is specified by $U_\theta(x) = (x/10)^2 + \theta^\T h(x) $, where $h(x)$ consists of ReLU basis functions
centered at the equi-spaced knots by $.1$ from $-4$ to $4$.
Figure \ref{fig:1d_example} presents the results from persistent training using MALA (see Appendix \ref{sec:append-Gaussian} for more details).
We observe the following patterns. (i) The replay-buffer data resemble the training data, including 3:1 proportions near the two modes $-2$ and $2$.
(ii) The long-run sample (MALA starting from real data) also resemble the training data.
(iii) The post-training sample (MALA starting from standard Gaussian noises) shows two local modes at $-2$ and $2$, but with proportions
different from 3:1 as in the training data.
(iv) The learned energy function exhibits two local modes about $-2$ and $2$, but is globally misaligned. The estimated energy at the local mode $-2$ is
substantially higher than at the other mode $2$, and hence is also higher than (for example) at $x=1$, an OOD point near no real data.
More troubling is that the estimated energies near $-2$ and $2$ may reverse the direction of relative magnitudes
from different training runs, as shown in Appendix \ref{sec:append-Gaussian}.
These results not only confirm the previous findings about long-run stability for persistent training \cite{nijkamp2020anatomy},
but also reveal a new phenomenon that the learned energies from persistent training
appear to be locally meaningful, but globally misaligned. Hence application of such learned energies to OOD detection is problematic.

Motivated by the 1D example, we provide some new theoretical understanding of persistent training of EBMs with
MCMC sampling which enables only local mixing for multimodal distributions with modes separated
by low-density (i.e., high-energy) barriers. Our discussion is heuristic, but highlights
the main ideas which can be exploited to develop formal analysis.
In the limit of the persistent training process (assumed to exist), let $\hat\theta$ be a limit value of the network parameter $\theta$, and
$\hat q$ be a limit distribution for the synthetic data (which can be represented by the empirical distribution of the replay buffer).
Then we expect that $(\hat \theta, \hat q)$ satisfy the following stationarity conditions: \vspace{-.1in}
\begin{itemize} \addtolength{\itemsep}{-.15in}
\item Sampling stationarity: $\hat q$ is invariant under the Markov transition $K_{\hat\theta}(\cdot,\cdot)$, i.e.,
\begin{align} \label{eq:invariant}
\textstyle{\int K_{\hat \theta} (x,\cdot) \hat q(x) \,\dif x = \hat q(\cdot)} .
\end{align}
\item Parameter stationarity: $\hat\theta$ is a stationary point of the expected gradient, i.e.,
\begin{align} \label{eq:stationary-ML}
 0 = -  \E_{\hat p} \Big\{\fpartial{}{\theta} U_{\hat\theta}(x) \Big\}
   + \E_{\hat q} \Big\{ \fpartial{}{\theta} U_{\hat\theta} (x) \Big\} .
\end{align}
\end{itemize} \vspace{-.05in}
Conversely, if the training process is initialized by any network parameter $\hat\theta$ and replay-buffer distribution $\hat q$ satisfying (\ref{eq:invariant})--(\ref{eq:stationary-ML}), then the network parameter and replay-buffer distribution are expected to stay as
$\hat\theta$ and $\hat q$ respectively during persistent training.
Therefore, any pair $(\hat\theta,\hat q)$ satisfying (\ref{eq:invariant})--(\ref{eq:stationary-ML}) can potentially be the limit values from the training process,
unless additional constraints are introduced.

From our numerical experiments (Sections \ref{sec:toy_example}--\ref{sec:mnist}) as well as the 1D example,
we postulate that persistent training involves the following mechanisms through (\ref{eq:invariant})--(\ref{eq:stationary-ML}).\vspace{-.1in}
\begin{itemize}\addtolength{\itemsep}{-.05in}
\item Eq.~(\ref{eq:invariant}) dictates the network parameter $\hat\theta$ such that the transition kernel $K_{\hat\theta}$ (depending on the
energy function $U_{\hat\theta}$) leaves the replay-buffer distribution $\hat q$ invariant.

\item Eq.~(\ref{eq:stationary-ML}) induces moment matching between $\hat p$ and $\hat q$ such that the replay-buffer distribution $\hat q$
resembles the real data distribution $\hat p$.
\end{itemize}\vspace{-.1in}
Note that if an MCMC sampler suffers local mixing for a target distribution with separated modes by low-density barriers,
then there is no unique invariant distribution.
In this setting, the relative weights between the modes may be arbitrary for samples from Langevin dynamics
as discussed in \citet{song2019denoising}, Section~3.2.2.
See Appendix \ref{sec:local-mixing} for further discussion.
To see how this sampling deficiency affects energy estimation, our interpretation of (\ref{eq:invariant})--(\ref{eq:stationary-ML}) proceeds as follows.
The network parameter $\hat\theta$ is learned such that the transition kernel $K_{\hat\theta}$ leaves invariant
the replay-buffer distribution $\hat q$ (which resembles the data distribution $\hat p$).
Given sufficient data and model capacity, it can be assumed that $\hat q \approx p^*$, where $p^*$ is the population version of $\hat p$.
Then $\hat\theta$ can be any parameter value such that
the global invariance holds approximately:
\begin{align} \label{eq:global-invariant}
\textstyle{\int K_{\hat \theta} (x,\cdot) p^*(x) \,\dif x \approx p^*(\cdot)} .
\end{align}
In Appendix \ref{sec:local-energy}, we show that if the transition kernel $K_{\hat\theta}$ enables only local mixing with low-density barriers in $p^*$,
then (\ref{eq:global-invariant}) can be satisfied non-uniquely, provided $U_{\hat\theta}$
is a {\it local energy function} which matches the (global) energy function for $p^*$
locally in separate regions up to possibly different additive constants.
Non-uniqueness of invariant distributions for sampling can be translated into non-uniqueness of energy functions learned.
An example of such local energy functions is the learned energy in Figure \ref{fig:1d_example}.
See Appendix \ref{sec:local-energy} for formal discussion of local energy functions,
and \citet{song2019denoising}, Section~3.2.1, for a related discussion about inaccurate score estimation in low-density regions.

In summary, we provide both numerical evidence and theoretical explanation for the phenomenon that in the presence of separated modes by
low-density barriers, persistent training using an MCMC sampler which suffers local mixing may only learn local energy functions,
which are locally meaningful but globally misaligned.
Stable long-run MCMC samples from real data
using such learned energy functions may be obtained, but this alone does not imply the (global) validity of the learned energies.

\subsection{Diffusion-assisted EBMs} \label{sec:DA-EBM}

From Section~\ref{sec:diagnosis}, we see that proper learning of energy functions for multimodal data
requires strategies to overcome local mixing in MCMC sampling
from multimodal model distributions. Better global sampling will likely make the learned energy function more globally aligned.
A direct approach is to tackle this problem only within the sampling operation by exploiting {\it enhanced sampling} techniques
such as serial tempering \cite{marinari1992, geyer1995annealing} and
parallel tempering \cite{Swendsen1986Replica, geyer1991markovml} with tempered distributions (see Appendix \ref{sec:enhanced-sampling}).
However, a drawback of this approach is that samples from the tempered distributions
do not contribute to updating network parameters (because no empirical data are modeled by the tempered distributions).

Alternatively, we realize that diffusion data created with multiple noise levels can also be used as auxiliary distributions
to bridge different modes in multimodal data or ``fill low-density regions'', as noted in \citet{song2019denoising} for score estimation.
We propose diffusion-assisted EBMs and develop an effective algorithm for ML learning.
Instead of using tempered distributions, our approach exploits diffusion data and their model distributions
for both sampling and parameter learning in an interdependent manner.
The diffusion data are used together with the original data to learn EBMs at multiple noise levels, and
the learned densities of diffusion data are then used, similarly as tempered distributions, to facilitate enhanced sampling.

First, we construct diffused real data as in \citet{ho2020ddpm}.
For an observation $z$ in the training set $\mathcal D$ and $t = 1,\ldots,T$, let
$z^{(t)} =  \sqrt{\alpha_t}\, z^{(t-1)} + \sqrt{1-\alpha_t}\, \varepsilon^{(t)}$, where $z^{(0)} = z$, $\varepsilon^{(t)} \sim \N(0,I)$ independently over $t$,
and $\alpha_1, \ldots, \alpha_{T} \in (0,1)$ are pre-specified noise variances such that $z^{(T)}$ is approximately distributed as $\N(0,I)$.
Equivalently, $z^{(t)}$ can be directly generated from $z$ as 
\begin{align}\label{eq:direct-diffusion}
    z^{(t)} \sim \N( \sqrt{\bar\alpha_t}\, z , (1-\bar\alpha_t)I ),
\end{align}
where $\bar\alpha_t = \prod_{j=1}^t \alpha_j$ for $t\ge 1$ and $\bar\alpha_0=1$.
We say that $z^{(t)}$ is a diffusion observation at time $t$ given $z$,
and the (marginal) distribution of $z^{(t)}$ is the diffusion data distribution at time $t$ if $z$ is randomly drawn from $\mathcal D$.

We propose to model the original data and diffusion data simultaneously through a joint EBM: \vspace{-.05in}
\begin{equation}\label{eq:DA-EBM}
    p_\theta(x,t) =\frac{ \exp(-U_\theta(x,t))}{ Z(\theta)},
\end{equation}
where $U_\theta(x,t)$ is an energy function in $(x,t)$ jointly and $Z(\theta) = \int \sum_{t=0}^T \exp(-U_\theta(x,t)) \dif x$.
The pair $(x,t)$ encodes that $x$ is treated as a diffusion observation at time $t$.
For any fixed $t$, the diffusion data at time $t$ are modeled by
the conditional distribution of $x$ given $t$ under (\ref{eq:DA-EBM}), which is an EBM with energy function $U_\theta(x,t)$ in $x$ only: \vspace{-.05in}
\begin{align} \label{eq:marginal-EBM}
p_\theta (x| t) = \frac{ \exp(-U_\theta(x,t))}{ Z_t(\theta)},
\end{align}
where $Z_t(\theta) = \int \exp(-U_\theta(x,t)) \dif x$.
In particular, $p_\theta(x|0)$ at time-$0$ represents an EBM for the original data,
whereas $p_\theta (x|T)$ at time-$T$ represents an EBM for a data distribution close to a Gaussian noise.
Hence model (\ref{eq:DA-EBM}) combines individual EBMs for original and diffusion data at different time steps
into a compact, joint model.

We pursue persistent training for the joint EBM (\ref{eq:DA-EBM}). Training Algorithm~\ref{alg:DA-EBM}
essentially follows ML-SA in Section~\ref{sec:background}.
Lines~3--6 are the sampling operation, and line~7 is the parameter updating operation.
However, we incorporate an enhanced sampling algorithm, called MALA within Gibbs
mixture sampling (MGMS), to achieve
joint sampling from the model distribution $p_\theta(x,t)$ in an efficient manner (Algorithm~\ref{alg:joint-sampling}).
In Algorithm~\ref{alg:joint-sampling}, lines~2--6 perform sampling
from the conditional distribution $p_\theta ( x| \tilde t_0)$ given the current time label $\tilde t_0$ using MALA,
starting from the current configuration $\tilde x_0$.
Line~7 performs exact sampling from the conditional distribution $p_\theta (t | \tilde x_1)$ given the new configuration $\tilde x_1$.
The same MGMS can also be used to implement post-training sampling, as shown in Algorithm~\ref{alg:post-sampling}.

The MGMS algorithm can be derived as an instance of labeled mixture sampling in \citet{tan2017mixturesampling}.
The two operations are called, respectively, Markov move and global jump, which draws $\tilde t_1$ given $\tilde x_1$ independently of $\tilde t_0$.
These two operations illustrate how MGMS may achieve proper sampling from a multimodal distribution:
moving through auxiliary distributions and jumping to different modes in the original distribution.
Moreover, our learning Algorithm~\ref{alg:DA-EBM} can be seen to extend self-adjusted mixture sampling in \citet{tan2017mixturesampling} for handling non-intercept parameters in learning EBMs. See Appendix \ref{sec:enhanced-sampling} for further discussion. 


We comment on some additional features of Algorithm~\ref{alg:joint-sampling}.
The Langevin step size $\sigma_t$ is allowed to depend on the time label $t$ to accommodate different variations in the diffusion data distributions.
We employ MALA instead of Langevin dynamics without rejection, to ensure sampling convergence for relatively large $\sigma_t$.
Moreover, the acceptance rates from the Metropolis--Hastings step can be used to adjust step sizes dynamically to automate tuning. See Appendix \ref{sec:app_images_impl}.

\begin{algorithm}[tb]
   \caption{ML learning for diffusion-assisted EBMs}\label{alg:DA-EBM}
\begin{algorithmic}[1]
   \STATE {\bfseries Input:} Real data $\mathcal{D}$, replay buffer $\mathcal{B}$, initial value $\theta$, learning rate $\gamma$, Langevin steps $L$, step sizes $\{\sigma_t\}_{t=0}^T$.
   \REPEAT

   \STATE {\bfseries Draw} $x_0$ from $\mathcal{D}$ and $t_1 \sim \text{Unif}({0, 1, \ldots, T })$;

   \STATE{\bfseries Generate} $x_1= x_0^{(t_1)}$ by (\ref{eq:direct-diffusion}) with $z=x_0$ and $t=t_1$;

    \STATE{\bfseries Draw} $(\tilde{x}_0, \tilde{t}_0)$ from $\mathcal{B}$;

    \STATE{\bfseries Sample} $(\tilde{x}_1, \tilde{t}_1)$ given $(\tilde{x}_0, \tilde{t}_0)$ by Algorithm \ref{alg:joint-sampling} and \\
    $\qquad$ reset $(\tilde{x}_0, \tilde{t}_0)$ to $(\tilde{x}_1, \tilde{t}_1)$ in $\mathcal{B}$;
    \STATE{\bfseries Update} using gradient ascent \\
    $\qquad$ $ \theta \leftarrow \theta + \gamma\{  -\fpartial{}{\theta} U_\theta(x_1, t_1) + \fpartial{}{\theta} U_\theta(\tilde{x}_1, \tilde{t}_1) \}$.
  \UNTIL{Converged}

\STATE {\bfseries Return:} ML estimator $\theta$.
 \end{algorithmic}
\end{algorithm}

\begin{algorithm}[tb]
   \caption{MALA within Gibbs mixture sampling}\label{alg:joint-sampling}
\begin{algorithmic}[1]
   \STATE {\bfseries Input:} Current observation $(\tilde{x}_0, \tilde{t}_0)$, Langevin steps $L$, step sizes $\{\sigma_t\}_{t=0}^T$.
   \FOR{$l=1$ {\bfseries to} $L$}

   \STATE {\bfseries Draw}  $\varepsilon \sim \N(0,I)$ and propose \\
   $\qquad$ $ \tilde x_{1/2} = \tilde x_0 - \frac{\sigma_{\tilde{t}_0}^2}{2} \nabla_x U_\theta(\tilde x_0, \tilde t_0) + \sigma_{\tilde{t}_0}\varepsilon$;

   \STATE{\bfseries Accept} $\tilde{x}_1 =\tilde{x}_{1/2}$ with probability $\min(1,r)$ or \\
   {\bfseries reject} $\tilde x_{1/2}$ and set $\tilde{x}_1=\tilde{x}_0$, where {\scriptsize $r=\exp( -H_1 + H_0)$}, \\
    {\scriptsize $H_1= U_\theta(\tilde x_{1/2}, \tilde t_0) + \frac{1}{2\sigma_{\tilde t_0}^2} \| \tilde x_0 - \tilde x_{1/2} + \frac{ \sigma_{\tilde t_0}^2}{2} \nabla_x U_\theta(\tilde x_{1/2}, \tilde t_0)\|^2_2$},\\
    {\scriptsize $H_0 = U_\theta(\tilde x_0, \tilde t_0) + \frac{1}{2\sigma_{\tilde t_0}^2} \| \tilde x_{1/2} - \tilde x_0 + \frac{ \sigma_{\tilde t_0}^2}{2 } \nabla_x U_\theta(\tilde x_0,\tilde t_0)\|^2_2$};\\

   \STATE{\bfseries Reset} $\tilde x_0=\tilde x_1$.

   \ENDFOR
    \STATE   {\bfseries Draw} $\tilde{t}_1$ given $\tilde{x}_1$ from Multinomial with class probabilities
    $p_\theta(t | \tilde x_1 ) = \frac{\exp(-U_\theta(\tilde x_1, t))}{\sum_{k=0}^T \exp(-U_\theta(\tilde x_1, k))} $.
    \STATE {\bfseries Return:} New observation $(\tilde{x}_1, \tilde{t}_1)$.
 \end{algorithmic} \vspace{-.05in}
\end{algorithm}

The joint EBM (\ref{eq:DA-EBM}) has the same form as in \citet{grathwohl2020JEM},
although a time step $t$ is involved instead of a class label $y$.
The two models are formulated for apparently different purposes. The training algorithms are also different in the initialization scheme,
persistent initialization here versus hybrid initialization in \citet{grathwohl2020JEM},
and in how MCMC sampling is handled.
\citet{grathwohl2020JEM} factorized the joint EBM density as $p_\theta(x,y) = p_\theta(y|x) p_\theta(x)$
and apply Langevin sampling (with a constant step size) to the marginal density $p_\theta(x)$.
This method is unsuitable in our setting, because the diffusion data exhibit different variations at different time steps.
See a related discussion of mixture importance sampling in Appendix \ref{sec:enhanced-sampling}.

\begin{algorithm}[tb]
   \caption{{Post-training sampling with MGMS }}\label{alg:post-sampling}
\begin{algorithmic}[1]
   \STATE {\bfseries Input:} Random noise and time label, $(\tilde{x}_0, \tilde{t}_0)=(\varepsilon, T)$, stopping criterion $M$.
   \REPEAT

    \STATE{\bfseries Sample} $(\tilde{x}_1, \tilde{t}_1)$ given $(\tilde{x}_0, \tilde{t}_0)$ by Algorithm \ref{alg:joint-sampling} and \\
    $\qquad$ reset $(\tilde{x}_0, \tilde{t}_0)$ to $(\tilde{x}_1, \tilde{t}_1)$;
  \UNTIL{$\tilde{t}_0$ reaches time $0$ for $M$ times}

\STATE {\bfseries Return:} Synthetic observation $\tilde{x}_0$ at time 0.
 \end{algorithmic}\vspace{-.05in}
\end{algorithm}  \vspace{-.05in}

\section{Related work} \label{sec:comparison}

There is a vast and growing literature on EBMs and related topics.
In addition to earlier discussions in Sections \ref{sec:introduction}--\ref{sec:proposed},
we discuss here  how our approach is compared with score-based diffusion modeling, mainly
the representative works \cite{sohl2015deep, song2019denoising, ho2020ddpm, song2021sde},
and with the diffusion recovery likelihood for EBM learning \cite{gao2020Diffusion}.

Our approach and the score-based approach differ in how to use these diffusion data for learning from the original data.
Denote as $p^*_t$ the population density of the diffusion data distribution at time $t$.
The score-based approach postulates a score-based model $s_\theta(x,t)$ for the score $\nabla_x \log p^*_t(x)$,
and then employs denoising score matching and extensions to implement training \cite{hyvrinen2005scorematch, vincent2011DSM}.
The log-density (or log-likelihood) $\log p^*_0(x)$ for the original data is not directly parameterized,
but can be approximated by solving the probability flow ODE based on the learned score \cite{song2021sde}.
From our experience, this method for likelihood calculation is computationally costly (see Appendix \ref{sec:app_images_impl}).
More importantly, the calculated likelihoods from the score-based approach are observed to
suffer the OOD reversal in our experiments on MNIST-type images.

By comparison, our approach involves EBM modeling and ML learning.
Hence the energy function (or the negative log-likelihood up to a constant) is analytically available for the learned EBMs.
The learned energy functions are globally more meaningful than obtained by previous methods, as shown by the superior
performance for OOD detection in our experiments.
The MGMS algorithm from our training algorithm can also be used
to generate new images from noises based on the learned EBMs (Algorithm~\ref{alg:post-sampling}).

For diffusion data as in \citet{ho2020ddpm},
the approach of \citet{gao2020Diffusion} postulates marginal EBMs in the form (\ref{eq:marginal-EBM}), rewritten as
$ p_\theta (x^{(t)} ) = \exp(-U_\theta( x^{(t)},t) ) / Z_t(\theta)$, and then derives conditional EBMs
$p_\theta(x^{(t-1)}|x^{(t)} )$,  where $x^{(t)}$ denotes a diffusion observation at time $t$.
The training algorithm of \citet{gao2020Diffusion} proceeds similarly as training marginal EBMs $p_\theta( x^{(t-1)})$, but
in each iteration
applies a fixed number of Langevin sampling steps
for $p_\theta(x^{(t-1)}|x^{(t)} )$, initialized by the diffusion observation $x^{(t)}$ (being conditioned on).
This is an instance of data initialization (or a variation of CD) as discussed in Section~\ref{sec:diagnosis}.
For post-training image generation, the same number of Langevin sampling steps are applied to sample from $p_\theta(x^{(t-1)}|x^{(t)} )$
with the initial value set to the final observation drawn at time $t$, iteratively from $t=T$ to $1$.
At time $T$, the initial value is a random noise.
This is reminiscent of image generation using learned short-run MCMC \cite{nijkamp2019shortrun},
although in a sequential manner as in backward diffusion.
The learned energy functions in the recovery likelihood approach appear to be globally misaligned and suffer the OOD reversal in
our experiments.

In the Appendix \ref{sec:reformulate-DRLK}, we show that the recovery likelihood approach can be equivalently formulated as training via CD ``bivariate" EBMs
for the pairs of observations $(x^{(t-1)}, x^{(t)})$, although it is originally presented in terms of training conditional EBMs
$p_\theta(x^{(t-1)}|x^{(t)} )$. Hence our approach differs from \citet{gao2020Diffusion} in maintaining persistent training while
combining marginal (univariate) EBMs into a joint EBM and incorporating enhanced MCMC sampling.
These choices help to improve the global alignment of the learned energy functions as seen in our experiments.

\section{Illustrative experiement} \label{sec:toy_example}

We provide a 2D example of four rings, where the data distribution exhibits multimodality and singularity.
Similar examples are reported in \citet{nijkamp2019shortrun} and \citet{gao2020Diffusion}.
We compare EBMs trained with different initialization schemes (Section~\ref{sec:background}), diffusion recovery likelihood (DRL) \cite{gao2020Diffusion}
and our method (DA-EBM).
See Appendix \ref{sec:append-four-ring} for further details of the experiment.

In Table \ref{tb:toy_example}, we plot learned energy functions (along the line $x_1=0$ and, for comparison,
anchored to have a minimum 0), long-run samples, and post-training samples.
An energy function is negative log-density up to a constant.
For DRL and DA-EBM, we plot learned energy functions at time 0 in Table \ref{tb:toy_example} and at other times in Appendix Table \ref{tb:toy_example_ctd}.

\begin{table}[!t] \vspace{-.1in}
\caption{Results from four-ring example}
\label{tb:toy_example}
\begin{center}
\begin{small}
\begin{sc}

\begin{tabular}{ c | c | c | c }
 \toprule[1pt]
  {\tiny \makecell{ Train. \\ Data \\ Info.}} &

     \multicolumn{3}{c}{    
        \begin{minipage}{.44\columnwidth}
     \includegraphics[width=\linewidth, height=18mm]{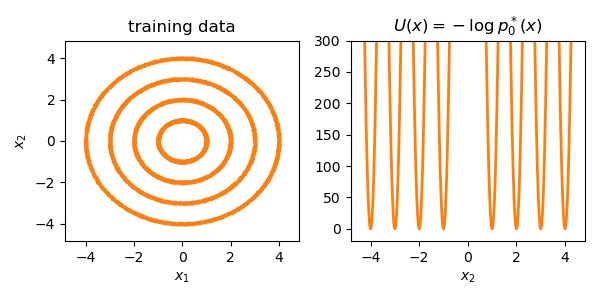}
    \end{minipage}
    }     
    \\

  \cmidrule[1pt](r){1-1}\cmidrule[1pt](l){2-2}\cmidrule[1pt](lr){3-3}\cmidrule[1pt](l){4-4}
     {\scriptsize Method} & {\scriptsize Energy}& {\scriptsize Long-run} & {\scriptsize Post}\\
   \cmidrule[1pt](r){1-1}\cmidrule[1pt](l){2-2}
      \cmidrule[1pt](r){1-1}\cmidrule[1pt](lr){2-2}\cmidrule[1pt](lr){3-3}\cmidrule[1pt](l){4-4}
      {\tiny  \makecell{ EBM \\CD}}
   &
        \begin{minipage}{.22\columnwidth}
         \includegraphics[width=\linewidth, height=18mm]{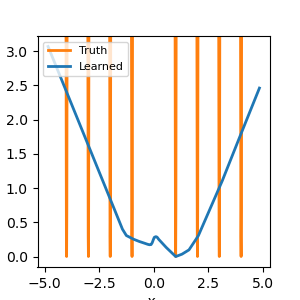}
        \end{minipage}
   &
        \begin{minipage}{.22\columnwidth}
         \includegraphics[width=\linewidth, height=18mm]{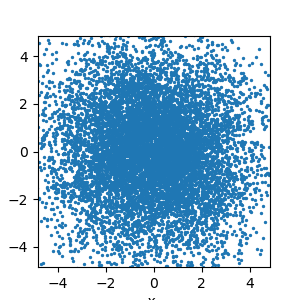}
        \end{minipage}
   &
        \begin{minipage}{.22\columnwidth}
         \includegraphics[width=\linewidth, height=18mm]{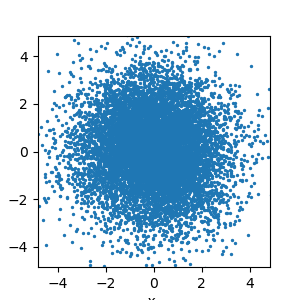}
        \end{minipage}\\ \cmidrule[.2pt]{1-4}

    {\tiny  \makecell{ EBM\\ PERS.}}
   &
        \begin{minipage}{.22\columnwidth}
         \includegraphics[width=\linewidth, height=18mm]{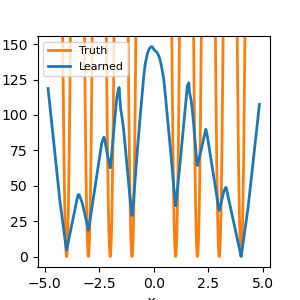}
        \end{minipage}
        &
        \begin{minipage}{.22\columnwidth}
         \includegraphics[width=\linewidth, height=18mm]{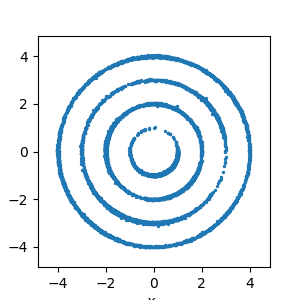}
        \end{minipage} 
        &
        \begin{minipage}{.22\columnwidth}
         \includegraphics[width=\linewidth, height=18mm]{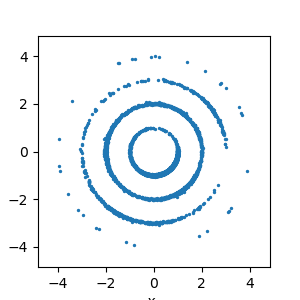}
        \end{minipage}\\ \cmidrule[.2pt]{1-4}
        
    {\tiny  \makecell{ 
    EBM \\NOISE}}
   &
        \begin{minipage}{.22\columnwidth}
         \includegraphics[width=\linewidth, height=18mm]{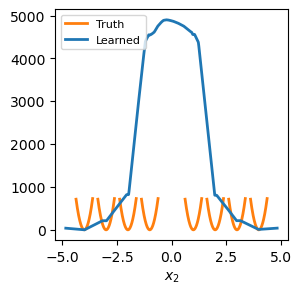}
        \end{minipage}
   &
        \begin{minipage}{.22\columnwidth}
         \includegraphics[width=\linewidth, height=18mm]{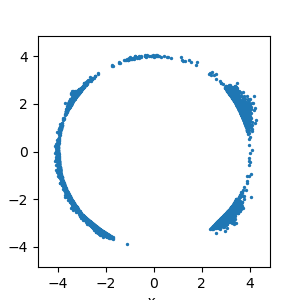}
        \end{minipage}
   &
        \begin{minipage}{.22\columnwidth}
         \includegraphics[width=\linewidth, height=18mm]{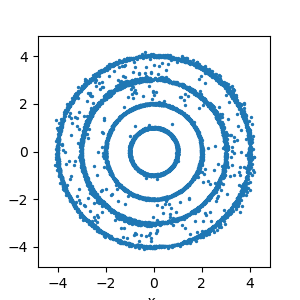}
        \end{minipage}\\ \cmidrule[.2pt]{1-4}

     {\tiny\makecell{
     EBM\\ HYBRID\\
     (5\% noise)} } &
        \begin{minipage}{.22\columnwidth}
         \includegraphics[width=\linewidth, height=18mm]{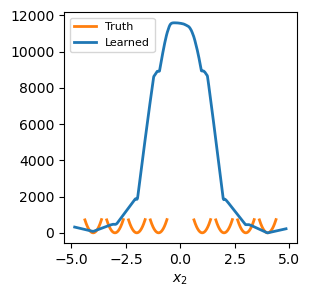}
        \end{minipage}
   &
        \begin{minipage}{.22\columnwidth}
         \includegraphics[width=\linewidth, height=18mm]{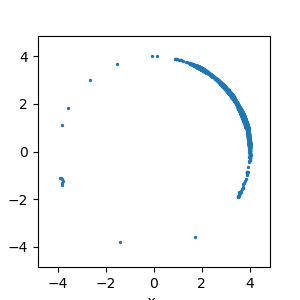}
        \end{minipage}
   &
        \begin{minipage}{.22\columnwidth}
         \includegraphics[width=\linewidth, height=18mm]{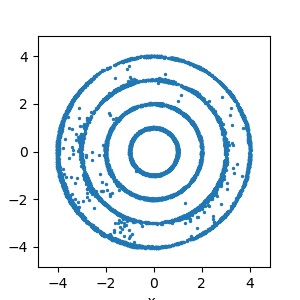}
        \end{minipage}\\ \cmidrule[.2pt]{1-4}

     {\tiny\makecell{
     DRL\\
     Data Init.} } &
        \begin{minipage}{.22\columnwidth}
         \includegraphics[width=\linewidth, height=18mm]{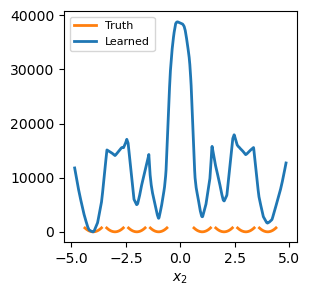}
        \end{minipage}
   &
        \begin{minipage}{.22\columnwidth}
         \includegraphics[width=\linewidth, height=18mm]{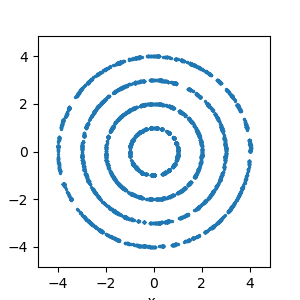}
        \end{minipage}
   &
        \begin{minipage}{.22\columnwidth}
         \includegraphics[width=\linewidth, height=18mm]{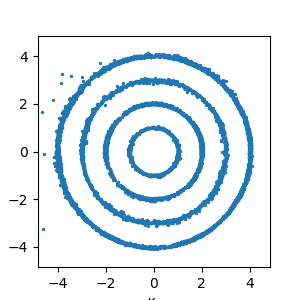}
        \end{minipage}\\ \cmidrule[.2pt]{1-4}
     {\tiny\makecell{
     DA-EBM\\
     Pers. Init.} } &
        \begin{minipage}{.22\columnwidth}
         \includegraphics[width=\linewidth, height=18mm]{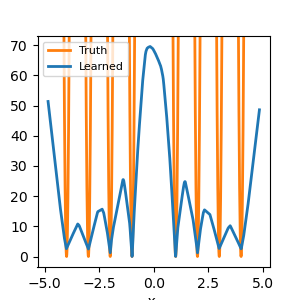}
        \end{minipage} &
        \begin{minipage}{.22\columnwidth}
         \includegraphics[width=\linewidth, height=18mm]{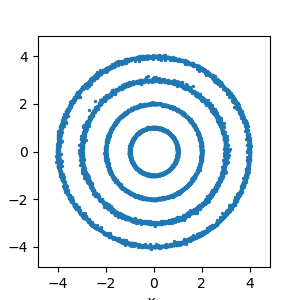}
        \end{minipage} &
        \begin{minipage}{.22\columnwidth}
         \includegraphics[width=\linewidth, height=18mm]{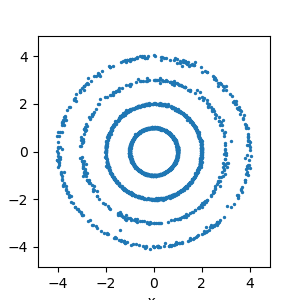}
        \end{minipage} \\ \bottomrule[1pt]
        
      
   \end{tabular}
\end{sc}
\end{small}
\end{center}
\vskip -0.2in
\end{table}

Table~\ref{tb:toy_example} confirms the properties of EBM methods in Table~\ref{tb:property_summary}.
(i) Among the first four rows for usual EBMs, only persistent training gives long-run stable samples.
For persistent training, the learned energy has local modes at the four rings, but is globally misaligned.
For example, the learned energy at $x_2=-2$ (where data are present) is incorrectly higher than that at $x_2 = -3.5$ (where data are absent).
Post-training sampling (from standard Gaussian noises) 
can barely reach the outmost ring.
(ii) The noise initialization (short-run MCMC) and hybrid initialization give similar results.
The learned energy functions have local modes at the rings, but the modes are almost invisible due to the large ranges of energy values,
which makes the gradient $\nabla_x U_\theta(x)$ strong to drive Langevin dynamics for data generation.
Post-training sampling can generate realistic samples, but long-run sampling is unstable.
(iii) Compared with noise and hybrid initializations, DRL also learns an energy function which has a large range and is globally misaligned.
The energy landscape has deeper local modes at the rings and hence long-run sampling can be stable.
(iv) Among all methods, our method gives an energy function which best approximates the truth.
The energy values are properly aligned across the local modes at the four rings.
Long-run sampling is stable. The post-sampling result shows that
our enhanced MCMC sampling can move across modes, with help from exploring all diffusion data distributions.

\begin{table}[!t] \vspace{-.1in}
\caption{MNIST results}
\label{tb:mnist_imgs}
\begin{center}
\begin{small}
\begin{sc}
     \begin{tabular}{ c | c | c  }
     \toprule[1pt]
     & \multicolumn{2}{c}{MNIST} 
        \\\hline
      \makecell{Train. Data \\
        \scriptsize (Long-run
        \\ \scriptsize starting \\
        \scriptsize points)
      }
      &      
      \multicolumn{2}{c}{    \begin{minipage}{.32\columnwidth}
         \includegraphics[width=\linewidth, height=21mm]{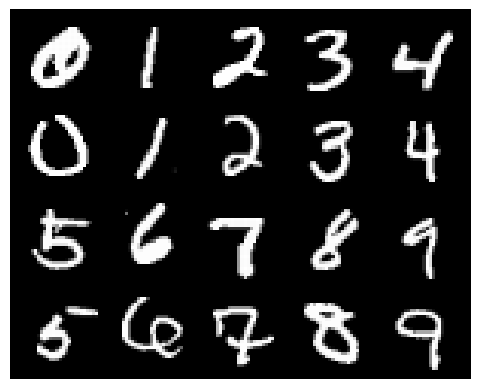}
        \end{minipage} }
        \\
      \cmidrule[1pt](r){1-1}\cmidrule[1pt](lr){2-2} \cmidrule[1pt](lr){3-3}
      Method &  Long-run & Post  \\
      \cmidrule[1pt](r){1-1}\cmidrule[1pt](lr){2-2} \cmidrule[1pt](lr){3-3} 
      {\small  \makecell{ EBM\\
    Pers. 
    }}   &
        \begin{minipage}{.32\columnwidth}
         \includegraphics[width=\linewidth, height=21mm]{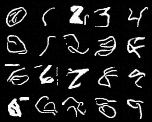}
        \end{minipage} &
        \begin{minipage}{.32\columnwidth}
         \includegraphics[width=\linewidth, height=21mm]{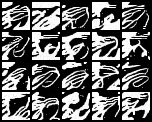}
        \end{minipage} 
        \\\cmidrule[.2pt]{1-3}
      {\small  \makecell{ EBM\\
    Hybrid \\
    (5\% noise) }}   &
        \begin{minipage}{.32\columnwidth}
         \includegraphics[width=\linewidth, height=21mm]{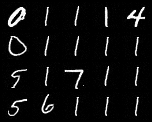}
        \end{minipage} &
        \begin{minipage}{.32\columnwidth}
         \includegraphics[width=\linewidth, height=21mm]{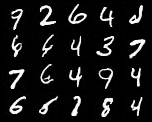}
        \end{minipage} 
        \\\cmidrule[.2pt]{1-3}
      {\small  \makecell{ DRL }}   &
        \begin{minipage}{.32\columnwidth}
         \includegraphics[width=\linewidth, height=21mm]{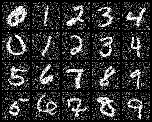}
        \end{minipage} &
        \begin{minipage}{.32\columnwidth}
         \includegraphics[width=\linewidth, height=21mm]{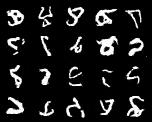}
        \end{minipage} 
        \\\cmidrule[.2pt]{1-3}
      {\small  \makecell{ DA-EBM }}   &
        \begin{minipage}{.32\columnwidth}
         \includegraphics[width=\linewidth, height=21mm]{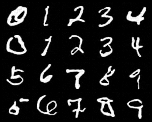}
        \end{minipage} &
        \begin{minipage}{.32\columnwidth}
         \includegraphics[width=\linewidth, height=21mm]{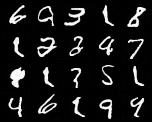}
        \end{minipage}  
        \\\bottomrule[1pt]
  \end{tabular}
\end{sc}
\end{small}
\end{center}
\vskip -0.2in
\end{table}

\section{Image experiments} \label{sec:mnist}

We study performances of several existing methods and ours on the two grayscale image datasets, MNIST and Fashion-MNIST.
In addition to EBM training methods as in Section~\ref{sec:toy_example}, we also
include Glow \cite{kingma2018glow} and continuous-time DDPM (cDDPM) \cite{song2021sde} when investigating OOD detection.
We do not report inception scores, because differences of the sampling results are visually clear between differen methods.
See Appendix \ref{sec:app_images_impl} for further details of the experiment.

\begin{table}[t]
\caption{OOD Results}
\label{tb:ood_hist_results}
\begin{center}
\begin{small}
\begin{sc}
\begin{tabular}{ c | c  | c| c }
\toprule[1pt]
     {\scriptsize Method} & {\scriptsize MNIST} & {\scriptsize EMNIST} & {\scriptsize KMNIST}  \\ \midrule

    {\tiny  \makecell{ EBM Pers. }}
   &
        \begin{minipage}{.22\columnwidth}
         \includegraphics[width=\columnwidth, height=18mm]{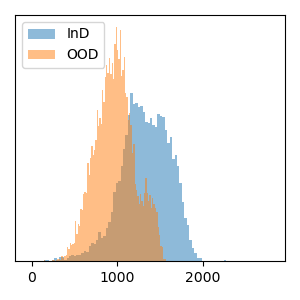}
        \end{minipage} & 
         \begin{minipage}{.22\columnwidth}
         \includegraphics[width=\columnwidth, height=18mm]{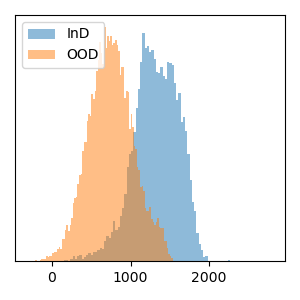}
        \end{minipage} & 
        \begin{minipage}{.22\columnwidth}
         \includegraphics[width=\columnwidth, height=18mm]{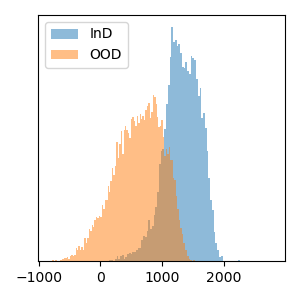}
        \end{minipage}\\ \cmidrule[.2pt]{1-4}

     {\tiny\makecell{
     EBM Hybrid \\
     (5\% noise)} } &
        \begin{minipage}{.22\columnwidth}
         \includegraphics[width=\columnwidth, height=18mm]{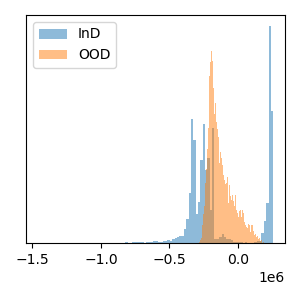}
        \end{minipage} & 
         \begin{minipage}{.22\columnwidth}
         \includegraphics[width=\columnwidth, height=18mm]{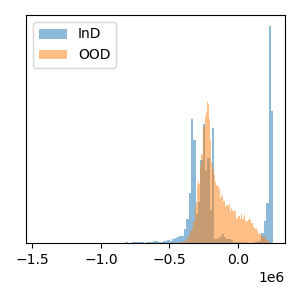}
        \end{minipage} & 
        \begin{minipage}{.22\columnwidth}
         \includegraphics[width=\columnwidth, height=18mm]{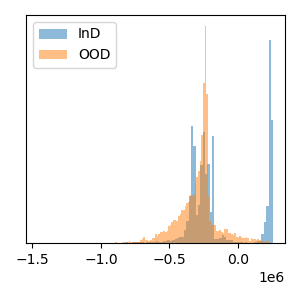}
        \end{minipage}\\ \cmidrule[.2pt]{1-4}       

     {\tiny\makecell{
     DRL} } &
        \begin{minipage}{.22\columnwidth}
         \includegraphics[width=\columnwidth, height=18mm]{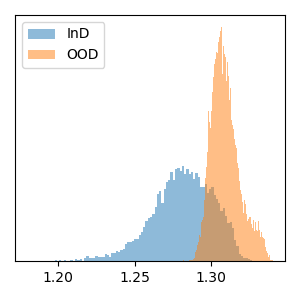}
        \end{minipage} & 
         \begin{minipage}{.22\columnwidth}
         \includegraphics[width=\columnwidth, height=18mm]{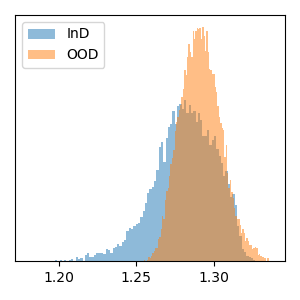}
        \end{minipage} & 
        \begin{minipage}{.22\columnwidth}
         \includegraphics[width=\columnwidth, height=18mm]{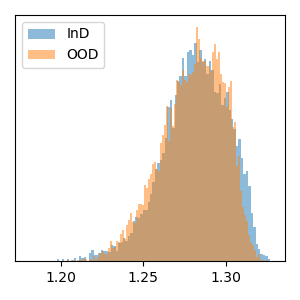}
        \end{minipage} \\ \cmidrule[.2pt]{1-4}       
     {\tiny\makecell{
     DA-EBM} } &
        \begin{minipage}{.22\columnwidth}
         \includegraphics[width=\columnwidth, height=18mm]{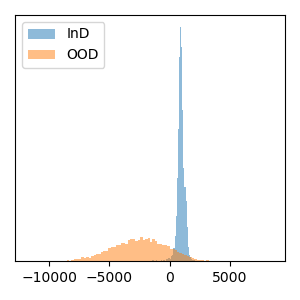}
        \end{minipage} & 
         \begin{minipage}{.22\columnwidth}
         \includegraphics[width=\columnwidth, height=18mm]{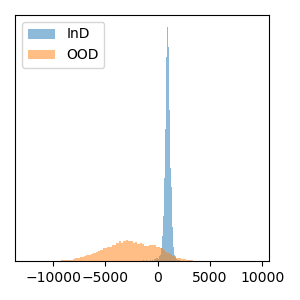}
        \end{minipage} & 
        \begin{minipage}{.22\columnwidth}
         \includegraphics[width=\columnwidth, height=18mm]{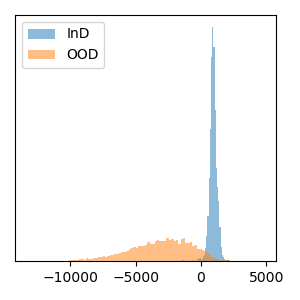}
        \end{minipage} \\ \cmidrule[.2pt]{1-4}
     {\tiny\makecell{
     GLOW} } &
        \begin{minipage}{.22\columnwidth}
         \includegraphics[width=\columnwidth, height=18mm]{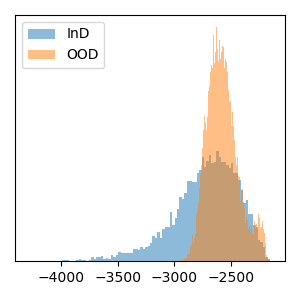}
        \end{minipage} & 
         \begin{minipage}{.22\columnwidth}
         \includegraphics[width=\columnwidth, height=18mm]{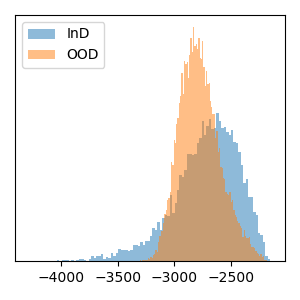}
        \end{minipage} & 
        \begin{minipage}{.22\columnwidth}
         \includegraphics[width=\columnwidth, height=18mm]{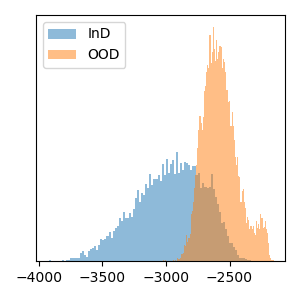}
        \end{minipage}\\ \cmidrule[.2pt]{1-4}
     {\tiny\makecell{
     cDDPM} } &
        \begin{minipage}{.22\columnwidth}
         \includegraphics[width=\columnwidth, height=18mm]{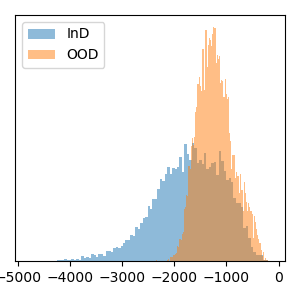}
        \end{minipage} & 
         \begin{minipage}{.22\columnwidth}
         \includegraphics[width=\columnwidth, height=18mm]{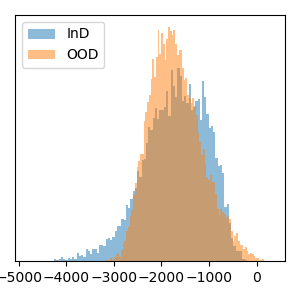}
        \end{minipage} & 
        \begin{minipage}{.22\columnwidth}
         \includegraphics[width=\columnwidth, height=18mm]{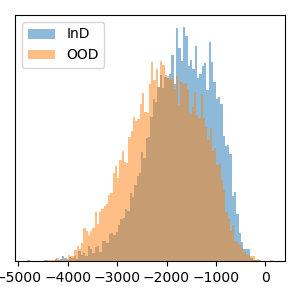}
        \end{minipage} \\  
       \bottomrule[1pt]
   \end{tabular}
\end{sc}
\end{small}
\end{center}
\vskip -0.4in
\end{table}

\subsection{Long-run stability and image generation}

Table \ref{tb:mnist_imgs} presents the results of long-run sampling and image generation on MNIST.
The {\it exact} starting points of long-run sampling are shown in the first row.
See Appendix \ref{sec:app_images_impl} for additional results, including from training on Fashion-MNIST or using the noise initialization method.

The long-run samples from hybrid-initialized EBM collapse into mostly digit 1 when the starting images differ from 1.
This is similar to the collapse of long-run samples to the outmost ring in Table \ref{tb:toy_example}.
The long-run samples from persistent-initialized EBM and DRL are stable, but
the former are seriously twisted whereas those from DRL have some arbitrary sprinkles added.
The long-run samples from DA-EBM remain close to the starting images.

For image generation, EBM with hybrid initialization seems to generate samples of good quality.
Persistently trained EBM cannot produce reasonable images.
The samples from DRL are twisted and differ from realistic digits.
The samples from DA-EBM, despite being persistently trained, are close to realistic digits.
Hence DA-EBM is the only method which performs satisfactorily in both
long-run stability (from real images) and image generation (from random noises).

\begin{table}[t]
\caption{OOD AUROC results}
\label{tb:ood_auc_results}
\vskip 0.15in
\begin{center}
\begin{small}
\begin{sc}

\begin{tabular}{ c | c c c   }\toprule
Method         &  Mnist &  EMnist & KMnist \\\midrule
EBM PERS.      &  0.83  &  0.92   &  0.93\\
EBM HYBRID     &  0.37  &  0.45   & 0.68 \\
DRL &  0.09  &   0.35  & 0.52 \\
DA-EBM         &  \textbf{0.93}  & \textbf{0.93} & \textbf{0.98} \\
GLOW           &  0.35  &  0.60   & 0.09 \\
cDDPM           &  0.27  &  0.52   & 0.63 \\
\bottomrule 
\end{tabular}%
\end{sc}
\end{small}
\end{center}
\vskip -0.1in
\end{table}

\subsection{Out-of-distribution detection}

For OOD detection, we use FashionMNIST as the in-distribution (InD) dataset and consider MNIST, EMNIST (Letters), and KMNIST (Japanese characters) as OOD datasets. All methods are trained on the training split of FashionMNIST, and OOD metrics are evaluated on the test split of each dataset. All methods use estimated log-densities
(i.e., log-likelihoods) as the scores to predict InD and OOD data.
Table~\ref{tb:ood_hist_results} presents the histograms of log-likelihoods, and Table \ref{tb:ood_auc_results} gives the AUROC (area under ROC) results.
Additional results can be found in Appendix \ref{sec:app_images_impl}.

Glow, cDDPM, and DRL all exhibit the OOD reversal. The learned models assign higher (or similar) likelihoods to the OOD data than to InD data.
For the hybrid-initialized EBM, the log-likelihoods of OOD data cover several modes of those of InD data.
This pattern was also seen in Table 2 of \citet{grathwohl2020JEM}.
In addition, the ranges of these log-likelihoods are of $10^6$ magnitude,
reminiscent of the unreasonable ranges of energy values in Table~\ref{tb:toy_example}.

Persistently trained EBM and DA-EBM are the only two methods which assign likelihoods in a proper direction (InD vs OOD).
But our learned likelihoods achieve better separation between InD and OOD data. The AUROC values from DA-EBM
not only improve upon the usual EBMs by margins at least 10\%, 1\%, and 5\% for the three OOD datasets, but also are much higher than those from Glow and cDDPM.
From the AUC-PR Table~\ref{tb:ood_aucpr_results} in Appendix, DA-EBM also outperforms (with over 10\% margins) the best in Table~1 of \citet{elflein2021out}
among EBM and other methods.

\section{Conclusion}

We propose diffusion-assisted EBMs and develop a persistent training algorithm.
The diffusion data are exploited for two benefits.
First, the diffusion data help to bridge different modes in the original data such that the density estimates between different modes can be appropriately aligned.
Second, the diffusion data also help to connect images and random noises to make post-training sampling through MCMC possible.
Our work opens the possibility of simultaneously achieving proper density estimation and post-training sampling.
See Appendix \ref{sec:append-limits} for a discussion about current limitations of our method.
It is desirable to further develop our method and conduct experiments on more complex image datasets and diverse tasks.


\clearpage


\bibliography{daebm_paper}
\bibliographystyle{icml2022}

\newpage
\appendix
\onecolumn

\section{Local energies} \label{sec:local-energy}

To formalize the discussion in Section~\ref{sec:diagnosis}, we introduce the following concepts.
For a (normalized) probability density $p(\cdot)$, say that $U(\cdot)$ is a {\it (global) energy function} for $p(\cdot)$ if
\begin{align*}
p (x) = \frac{\exp(-U(x))}{ \int  \exp(-U(x^\prime))\,\dif x^\prime} .
\end{align*}
Hence an energy function usually defined is a global energy function.
Suppose that the support of $p(\cdot)$, $\{x: p(x)>0\}$, is arbitrarily partitioned into 2 regions $B_1$ and $B_2$.
Say that $\tilde U(\cdot)$ is a {\it local energy function} for $p(\cdot)$ with respect to $(B_1,B_2)$ if
\begin{align}
p (x) = \sum_{j=1}^2  p(B_j) \frac{\exp(-\tilde U(x))}{ \int_{B_j} \exp(-\tilde U(x^\prime))\,\dif x^\prime} 1_{B_j}(x), \label{eq:local-energy}
\end{align}
where $p(B_j) = \int_{B_j} p(x)\,\dif x$ and $1_{B_j}$ is the indicator function for $B_j$.
Informally, a local energy function can be locally normalized and then properly combined into a given probability distribution.
This definition suffices in the following discussion of multimodal distributions with modes separated by zero-density barriers.
For nonzero- but low-density barriers, our discussion can be extended with more technical complexity, for example, allowing
equality (\ref{eq:local-energy}) to hold approximately, as measured by some distance between the two sides of (\ref{eq:local-energy}).

Let $p_0$ be a mixture of two disjoint densities,
\begin{align}
    p_0(x) = \pi p_{01}(x) + (1-\pi) p_{02}(x), \label{eq:mixture}
\end{align}
where $p_{01}$ and $p_{02}$ are two disjoint (normalized) densities with supports $B_1$ and $B_2$ respectively
($B_1 \cap B_2= \emptyset$) and $\pi\in(0,1)$ is the relative weight for $p_{01}$.
Then there exist an infinite collection of local energy functions with respect to $(B_1, B_2)$,
such that none of them is a global energy function for $p_0(\cdot)$.
These local energy functions can be defined as
$ (-\log p_{01} +c_{01}) 1_{B_1} + (-\log p_{02} + c_{02}) 1_{B_2} + \infty 1_{(B_1\cup B_2)^c}$ for arbitrary constants $(c_{01},c_{02})$.

For a multimodal distribution with strictly separated modes,
a local energy function (not necessarily a global energy function) can be used to achieve global invariance (\ref{eq:mix-invariant}) below,
with an MCMC sampler such as random walk Metropolis sampling or
Langevin sampling, which only enables local moves near individual modes.
Let $\tilde U(\cdot)$ be a local energy function for $p_0(\cdot)$ as above,
and let $K_{\tilde U} (\cdot,\cdot)$ be the transition kernel of an MCMC sampler, depending on $\tilde U(\cdot)$, such that (see Appendix \ref{sec:local-mixing})
\begin{align}
& K_{\tilde U} (x, x^\prime) =0 \text{ for } x\in B_1, x^\prime\in B_2 \text{ or } x\in B_2, x^\prime \in B_1, \label{eq:local-sample1} \\
& \textstyle{ \int_{B_j} K_{\tilde U} ( x, x^\prime) p_{0j}(x) \,\dif x = p_{0j}( x^\prime) } \text{ for } x^\prime \in B_j,j=1,2.  \label{eq:local-sample2}
\end{align}
Informally, the sampler only moves points within $B_1$ or $B_2$, and leaves $p_{01}$ or $p_{02}$ (locally) invariant on $B_1$ or $B_2$.
Then the global invariance holds (see the proof later):
\begin{align} \label{eq:mix-invariant}
\textstyle{ \int K_{\tilde U} ( x, x^\prime) p_0 (x) \,\dif x = p_0 ( x^\prime) }, \; x^\prime \in B_1\cup B_2,
\end{align}
that is, the sampler leaves the mixture distribution $p_0$ invariant on $B_1\cup B_2$.
Note that property (\ref{eq:mix-invariant}) does not imply that the Markov chain with kernel $K_{\tilde U}$ and any initial point
converges to the target distribution $p_0$; the chain is restricted to $B_j$ if the initial point is from $B_j$.
But a key message here is that in the presence of separated modes, global invariance (\ref{eq:mix-invariant}) can be {\it non-uniquely} achieved by
using any local energy function $U$ with an MCMC sampler which enables only local mixing.

We give a direct proof of Eq.~(\ref{eq:mix-invariant}). For $x^\prime \in B_1$,
the left-hand side of (\ref{eq:mix-invariant}) can be calculated as
\begin{align*}
\int K_{\tilde U} ( x, x^\prime) p_0 (x) \,\dif x
= \int_{B_1} K_{\tilde U} ( x, x^\prime) p_0 (x) \,\dif x
= \pi \int_{B_1} K_{\tilde U} ( x, x^\prime) p_{01} (x) \,\dif x
= \pi p_{01} (x^\prime),
\end{align*}
where the first equality is from (\ref{eq:local-sample1}), the second from (\ref{eq:mixture}) and $B_1 \cap B_2= \emptyset$,
and the third from (\ref{eq:local-sample2}).
Similarly, for $x^\prime \in B_2$, the left-hand side of (\ref{eq:mix-invariant}) can be shown to be $(1-\pi) p_{02} (x^\prime)$.
Combining the two cases gives the desired equality (\ref{eq:mix-invariant}), due to (\ref{eq:mixture}) and $B_1 \cap B_2= \emptyset$.

\section{Local mixing} \label{sec:local-mixing}

We discuss the local mixing behavior of Langevin sampling in the presence of separated modes.
For simplicity, we focus on the situation where two modes are strictly separated as two disjoint components of a mixture as in Appendix \ref{sec:local-energy}.
The main point can also be extended to separated modes by nonzero- but low-density barriers.

Consider a mixture of two disjoint densities as defined in (\ref{eq:mixture}).
Let $U_0(x) = -\log p_0(x) + c_0$ and $U_{0j}(x) = -\log p_{0j}(x) + c_{0j}$ for $j=1,2$, where
$c_0, c_{01}, c_{02}$ are three arbitrary (unknown) constants.
Then the gradient of the energy function $U_0$ is
\begin{align} \label{eq:mix-U}
    \nabla_x U_0(x) =
    \begin{cases}
    \nabla_x U_{01}(x), & x \in B_1, \\
    \nabla_x U_{02}(x), & x \in B_2. \\
    \end{cases}
\end{align}
From this relationship, if an initial point is from the support of $p_{01}$, then Langevin sampling treats $p_{01}$ as the target (invariant) distribution. In this case,
the Langevin chain only travels within the support of $p_{01}$.
Similarly, if an initial point is from the support of $p_{02}$, then the Langevin chain only travels within the support of $p_{02}$.
Therefore, cross-mode traveling is impossible (or almost impossible with low-density barriers), and
the Langevin chain is always trapped in the region where the initial point is located.
From another perspective, the Langevin chain using the energy function $U_0$
may admit a mixture of $p_{01}$ and $p_{02}$ with {\it any} relative weight, not just $p_0$, as an invariant distribution.
The relative weight information is lost for Langevin sampling.
See \citet{song2019denoising}, Section~3.2.2., for a related discussion.

As discussed in Appendix \ref{sec:local-energy},
a local energy function for $p_0$ can be defined as
$\tilde U(x) = U_{01}(x)$ if $x \in B_1$ or $U_{02}(x)$ if $x \in B_2$ or $\infty$ otherwise.
From the gradient expression (\ref{eq:mix-U}), Langevin sampling using $\tilde U$ as the energy function is indistinguishable from Langevin sampling
using the (global) energy function $U_0$. This also explains why the global invariance (\ref{eq:mix-invariant})
is satisfied for Langevin sampling using any local energy function in the presence of separated modes.

\section{Enhanced sampling} \label{sec:enhanced-sampling}

To overcome local mixing for multimodal distributions,
various enhanced sampling algorithms can be used while introducing auxiliary distributions as mentioned in Section \ref{sec:DA-EBM}.
For a distribution with energy function $U$, a popular scheme in practice as well as theoretical studies
\cite{ge2018tempering, chen2019accelerating} is to introduce tempered distributions with energies
$\{\beta_j U: j=1,\ldots,J\}$ for a decreasing sequence of inverse temperatures $1=\beta_1 > \cdots > \beta_J \approx 0$.
The Markov chains at higher temperatures (smaller $\beta_j$) may travel more easily between modes,
which then provide ``bridges" for the Markov chains at lower temperatures to explore those modes (even separated by low-density barriers).
More generally, a heuristic guideline is that those auxiliary distributions should be easy to sample from and
connected (or overlapped) with the original distribution, so as to help the sampling algorithm to explore the original distribution.

Return to the mixture distribution of two disjoint densities, as defined in (\ref{eq:mixture}).
In this case, introducing tempered distributions does not work because each tempered distribution remains a mixture of two disjoint densities.
Alternatively, we may introduce another auxiliary distribution to fill the low-density regions in $p_0$.
Let $p_1(x)$ be an ``umbrella'' unimodal density such that its support $\{x: p_1(x)>0\}$ contains $B_1 \cup B_2$,
and let $U_1(x) = -\log p_1(x) + c_1$, where $c_1$ is an arbitrary (unknown) constant.
The difference of log-normalizing constants (or free energy difference) between $U_0$ and $U_1$ is $\delta=c_1-c_0$.
To see advantages of our sampling Algorithm \ref{alg:joint-sampling} and a connection to our learning Algorithm \ref{alg:DA-EBM},
we compare several sampling algorithms in the special case of only one auxiliary distribution.

\textbf{Simple importance sampling (SIS).}
The first is SIS with the design density taken to be $p_1$.
By unimodality of $p_1$, we assume that Langevin sampling from $p_1$ achieves global (fast) mixing.
\begin{itemize}
\item[(i)] Draw a sample $\{ \tilde x_i\}_{i=1}^n$ from $p_1$ using MALA from an initial value $\tilde x_0$.
\item[(ii)] Draw a sample of some size $k$, $\{i_1, \ldots, i_k\}$, with replacement from $\{1,2,\ldots,n\}$ with the probability of selecting $i$ equal to
\begin{align*}
w_i = \frac{ (p_0/p_1)(\tilde x_i) }{\sum_{l=1}^n (p_0/p_1)(\tilde x_l) }
= \frac{ \me^{-U_0(\tilde x_i)+U_1(\tilde x_i)} }{ \sum_{l=1}^n \me^{-U_0(\tilde x_l)+U_1(\tilde x_l)} }.
\end{align*}
\end{itemize}
Then $\{ \tilde x_{i_1}, \ldots, \tilde x_{i_k}\}$ is a valid sample from $p_0$, in the sense that
sample averages are consistent for the corresponding expectations under $p_0$ as $n\to\infty$.
The weights $w_i$ can be calculated without knowing $\delta$,
but the sample is only approximately unbiased for large $n$.
The algorithm is non-iterative because increasing $n$ requires redrawing indices from $\{1,\ldots,n\}$.
For SIS to perform properly, the ratio $p_0(x)/p_1(x) \propto \me^{-U_0(x)+U_1(x)}$ need to have a finite variance under $p_1$,
which can be difficult to assess in practice.

\textbf{Mixture importance sampling (MIS).}
The second is MIS with the design density defined by (for example) the energy function
$U_\bullet (x) = -\log ( \me^{-U_0} + \me^{-U_1} )$. The associated density function is a mixture of $p_0$ and $p_1$,
\begin{align*}
p_\bullet (x) \propto p_0(x) + \me^{-\delta} p_1(x),
\end{align*}
where the relative weight for $p_0$ is $1/(1+\me^{-\delta})$ depending on $\delta$.
The mixture density $p_\bullet$ may be multimodal, but the two regions $B_1$ and $B_2$
are ``connected'' under $p_\bullet$ through probability mass from $p_1$ instead of completely separated in $p_0$,
so that Langevin sampling from $p_\bullet$ may achieve global mixing.
Given an initial value $\tilde x_0$, the MIS algorithm iterates for $i=1,\ldots,n$ as follows.

[Alternatively, the design density can be defined as $p_\bullet (x) = \frac{1}{2} p_0(x) + \frac{1}{2} p_1(x)$,
which has (equal) relative weights independent of $\delta$. But the associated energy function
$U_\bullet (x) = -\log ( \me^{-U_0} + \me^{-U_1 + \delta})$ depends on $\delta$.
An MIS algorithm similar as below can be derived, but both steps (i)--(ii) requiring the value of $\delta$.]

\begin{itemize}
\item[(i)] Draw $\tilde x_i$ using MALA transition for target distribution $p_\bullet$ (with energy $U_\bullet$) from initial value $\tilde x_{i-1}$.
\item[(ii)] Draw $\tilde s_i =1$ or $2$ with probability $v(\tilde x_i)$ or $1-v(\tilde x_i)$ respectively, where
\begin{align*}
& v(x) = \frac{ \frac{1}{1+\me^{-\delta}} p_0(x) }{p_\bullet(x)} = \frac{ \me^{-U_0(x)} }{ \me^{- U_0(x)} + \me^{-U_1(x)} },
\end{align*}
which is independent of $\delta$.
\end{itemize}
Then $\{ \tilde x_i: \tilde s_i=1, i=1,\ldots,n\}$ is a valid sample from $p_0$.
Compared with SIS, the weight $v(x)$ automatically has a finite variance under $p_\bullet$, because
\begin{align*}
\quad \int \left\{ \frac{ p_0(x) }{ p_\bullet(x)} \right\}^2 p_\bullet(x)\,\dif x
= (1+\me^{-\delta})\int \frac{ p^2_0(x) }{p_0(x) + \me^{-\delta} p_1(x) } \,\dif x \le 1+\me^{-\delta}.
\end{align*}
Hence MIS can be more stable than SIS.
On the other hand, the effectiveness of MALA sampling from the mixture $p_\bullet$ may be limited
because $p_0$ and $p_1$ are spatially of different scales and hence
different Langevin step sizes are desired for traversing the regions of $p_0$ and $p_1$.

\textbf{Metropolis within Gibbs mixture sampling (MGMS).}
The third is MGMS or, more specifically, MALA within Gibbs mixture sampling, which constitutes a special case
of our sampling Algorithm \ref{alg:joint-sampling} with only one auxiliary distribution. Consider the joint distribution
\begin{align*}
p(x,s) =
    \begin{cases}
    \frac{1}{1+\me^{-\delta}} p_0(x), & s=1, \\
    \frac{1}{1+\me^\delta} p_1(x), & s=2, \\
    \end{cases}
\end{align*}
which is called a labeled mixture \cite{tan2017mixturesampling} because each point $x$ is paired with a label $s$.  The marginal density of $x$ under $p(x,s)$ is $p_\bullet(x)$.
As an energy function of $p(x,s)$, let
\begin{align*}
U(x,s) =
    \begin{cases}
    U_0(x), & s=1, \\
    U_1(x), & s=2. \\
    \end{cases}
\end{align*}
Given initial values $(\tilde x_0, \tilde s_0)$, the MGMS algorithm iterates for $i=1,\ldots,n$ as follows.
\begin{itemize}
\item[(i)] If $\tilde s_{i-1}=1$, draw $\tilde x_i$ using MALA transition for target distribution $p_0$ (with energy $U_0$) from initial value $\tilde x_{i-1}$.
If $\tilde s_{i-1}=2$, draw $\tilde x_i$ using MALA transition for target distribution $p_1$ (with energy $U_1$) from initial value $\tilde x_{i-1}$.
\item[(ii)] Draw $\tilde s_i \in \{1,2\}$ as in step (ii) of MIS.
\end{itemize}
Then $\{ \tilde x_i: \tilde s_i=1, i=1,\ldots,n\}$ is a valid sample from $p_0$.
The two steps of MGMS perform sampling from the conditional distribution $p(x|s= \tilde s_{i-1})$ using MALA and then
sampling exactly from $p(s | x= \tilde x_i)$, both under the joint distribution $p(x,s)$.
These two steps are called Markov move and global jump, which draws $\tilde s_i$ independently of $\tilde s_{i-1}$ \cite{tan2017mixturesampling}.
[Alternatively, a local-jump scheme can be used for drawing $\tilde s_i$ by Metropolis--Hastings sampling from $p(s | x= \tilde x_i)$ with initial value $\tilde s_{i-1}$;
this is known as serial tempering \cite{marinari1992, geyer1995annealing}.]
Compared with MIS, the algorithm allows MALA sampling of $x$ with different step sizes from $p_0$ and $p_1$ in a principled manner.

The performance of MGMS (as well as MIS) depends on the (unknown) value of $\delta$ or equivalently
the relative weight for $p_0$, i.e., $p(s=1)= 1/(1+\me^{-\delta})$.
If $\delta$ is too negative (tending to $-\infty$), then the relative weight for $p_0$ is too large (making $p_\bullet$ close to $p_0$
and Langevin sampling suffer local mixing).
If $\delta$ is too positive (tending to $\infty$), then the relative weight for $p_0$ is too small (making the realixed sample size for $p_0$
small even for a large $n$).

\textbf{Self-adjusted mixture sampling (SAMS).}
To address the preceding issue in MGMS, an adaptive approach is to iteratively shift the energy functions $U_0$ and $U_1$ to
$U_0 + \zeta_0$ and $U_1 + \zeta_1$ for some additive constants $\zeta_0$ and $\zeta_1$ (while fixing initial $U_0$ and $U_1$),
such that $\delta=0$ or $p(s=1)=p(s=2)=1/2$ based on the new energy functions.
An SA algorithm using MGMS in the sampling step leads to SAMS \cite{tan2017mixturesampling}.
[Alternatively, a local-jump instead of global-jump scheme can be used in the sampling step \cite{liang_etal_2007}.]
Given initial values $(\tilde x_0, \tilde s_0)$ and $\zeta_0^{(0)}=\zeta_1^{(0)}=0$,
the SAMS algorithm iterates for $i=1,\ldots,n$ as follows.
\begin{itemize}
\item[(i)] If $\tilde s_{i-1}=1$, draw $\tilde x_i$ using MALA with energy $U_0 + \zeta_0^{(i-1)}$ from initial value $\tilde x_{i-1}$.
If $\tilde s_{i-1}=2$, draw $\tilde x_i$ using MALA with energy $U_1 + \zeta_1^{(i-1)}$ from initial value $\tilde x_{i-1}$.
\item[(ii)] Draw $\tilde s_i =1$ or $2$ with probability $v(\tilde x_i)$ or $1-v(\tilde x_i)$ respectively, where
\begin{align*}
& v(x) = \frac{ \me^{-U_0(x)-\zeta_0^{(i-1)}} }{ \me^{- U_0(x)-\zeta_0^{(i-1)}} + \me^{-U_1(x)-\zeta_1^{(i-1)}} },
\end{align*}
\item[(iii)] Update
\begin{align*}
& \zeta_0^{(i)} = \zeta_0^{(i-1)} + \gamma ( -\frac{1}{2} + 1\{\tilde s_i=1\} ),\\
& \zeta_1^{(i)} = \zeta_1^{(i-1)} + \gamma ( -\frac{1}{2} + 1\{\tilde s_i=2\} ),
\end{align*}
where $\gamma$ is a learning rate.
\end{itemize}
Remarkably, it can be verified that the SAMS algorithm is equivalent to our learning Algorithm \ref{alg:DA-EBM}
in the special case of learning the ``intercept" parameter $\zeta= (\zeta_0,\zeta_1)$, where the model energy function is parameterized as
\begin{align*}
U_\zeta(x,s) =
    \begin{cases}
    U_0(x) +\zeta_0, & s=1, \\
    U_1(x) +\zeta_1, & s=2, \\
    \end{cases}
\end{align*}
with fixed initial energies $(U_0, U_1)$, and the real-data distribution of $(x,s)$ is assumed such that
the marginal probabilities of $s=1$ and 2 are both $\frac{1}{2}$.
In this sense, the proposed Algorithm \ref{alg:DA-EBM} can be seen to extend SAMS for learning non-intercept parameters in EBMs.

\section{Reformulation of diffusion recovery likelihood} \label{sec:reformulate-DRLK}

We provide a reformulation of the recovery likelihood approach of \citet{gao2020Diffusion}, as mentioned in Section~\ref{sec:comparison}.
For simplicity, we consider only a single pair of observation $(x^{(t-1)}, x^{(t)})$, satisfying
$x^{(t)} = \sqrt{\alpha_t} x^{(t-1)} + \sqrt{1-\alpha_t} \varepsilon^{(t)}$, where $\varepsilon^{(t)} \sim \N(0,I)$.
Assume that $x^{(t-1)}$ satisfies an EBM with energy function $U_\theta (x^{(t-1)},t-1)$, i.e.,
\begin{align*}
p_\theta ( x^{(t-1)} ) \propto \exp( - U_\theta ( x^{(t-1)},t-1)  ).
\end{align*}
Then $x^{(t-1)}$ given $x^{(t)}$ satisfies a conditional EBM, with the conditional density
\begin{align*}
p_\theta( x^{(t-1)} | x^{(t)} ) \propto  \exp \left\{ -U_\theta(x^{(t-1)}, t-1)-\frac{\|x^{(t)} - \sqrt{\alpha_{t}} x^{(t-1)}\|_2^2 }{2 (1-\alpha_{t}) }  \right\} ,
\end{align*}
up to a multiplicative constant, free of $x^{(t-1)}$ but depending on $x^{(t)}$.
The gradient of $\log p_\theta( x^{(t-1)} | x^{(t)} ) $ can be shown to be
\begin{align} \label{eq:DRLK-grad}
\frac{\partial}{\partial\theta} \log p_\theta( x^{(t-1)} | x^{(t)} ) = -\frac{\partial}{\partial\theta} U_\theta (x^{(t-1)}, t-1)
+ \E_{p_\theta (z^{(t-1)}|x^{(t)} )} \left\{ \frac{\partial}{\partial\theta} U_\theta (z^{(t-1)}, t-1) \right\} ,
\end{align}
where $\E_{p_\theta (z^{(t-1)}|x^{(t)} )} (\cdot)$ denotes the (conditional) expectation over $z^{(t-1)}$ with respect to $p_\theta (z^{(t-1)}|x^{(t)} )$.
Note that $x^{(t-1)}$ is the data point observed together with $x^{(t)}$, and $z^{(t-1)}$ is a general point.
In the recovery likelihood approach, the gradient (\ref{eq:DRLK-grad}) is approximated by
\begin{align}\label{eq:DRLK-grad-approx}
-\frac{\partial}{\partial\theta} U_\theta (x^{(t-1)}, t-1)
+ \frac{\partial}{\partial\theta} U_\theta (\tilde x^{(t-1)}, t-1)  ,
\end{align}
where $\tilde x^{(t-1)}$ is drawn by running (for example) $L$ Langevin sampling steps targeting the conditional density $p_\theta (z^{(t-1)}|x^{(t)} )$, with the initial value set to $x^{(t)}$.

For the reformulation, we note that the pair $(x^{(t-1)}, x^{(t)})$ satisfies a ``bivariate" EBM, with the joint density
\begin{align*}
p_\theta( x^{(t-1)}, x^{(t)} ) \propto  \exp \left\{ -U_\theta(x^{(t-1)}, t-1)-\frac{\|x^{(t)} - \sqrt{\alpha_{t}} x^{(t-1)}\|_2^2 }{2 (1-\alpha_{t}) }  \right\} ,
\end{align*}
up to a multiplicative constant, free of $(x^{(t-1)}, x^{(t)})$.
The ``bivariate" EBM $p_\theta( x^{(t-1)}, x^{(t)} )$ is more fundamental than the conditional EBM $p_\theta( x^{(t-1)} | x^{(t)} )$,
because the latter is derived from the former, but not conversely.
The gradient of $\log p_\theta( x^{(t-1)}, x^{(t)} ) $ can be shown to be
\begin{align} \label{eq:bivariate-grad}
\frac{\partial}{\partial\theta} \log p_\theta( x^{(t-1)}, x^{(t)} ) = -\frac{\partial}{\partial\theta} U_\theta (x^{(t-1)}, t-1)
+ \E_{p_\theta (z^{(t-1)}, z^{(t)} )} \left\{ \frac{\partial}{\partial\theta} U_\theta (z^{(t-1)}, t-1) \right\} ,
\end{align}
where $\E_{p_\theta (z^{(t-1)},z^{(t)} )} (\cdot)$ denotes the expectation with respect to $p_\theta (z^{(t-1)}, z^{(t)} )$.
Note that $( x^{(t-1)}, x^{(t)} )$ is the observed pair, and $(z^{(t-1)},z^{(t)} )$ is a general pair of points.
Consider training the bivariate EBM by CD (contrastive divergence). The expectation term in (\ref{eq:bivariate-grad}) can be approximated by
\begin{align}\label{eq:bivariate-grad-approx}
-\frac{\partial}{\partial\theta} U_\theta (x^{(t-1)}, t-1)
+ \frac{\partial}{\partial\theta} U_\theta (\tilde x^{(t-1)}, t-1)  ,
\end{align}
where $(\tilde x^{(t-1)}, \tilde x^{(t)})$ is drawn from an Markov transition kernel targeting  $p_\theta (z^{(t-1)}, z^{(t)} )$, with the initial value set
to the observed data point $(x^{(t-1)}, x^{(t)})$. Suppose that $(\tilde x^{(t-1)}, \tilde x^{(t)})$  is drawn as follows, given the initial value $(x^{(t-1)}, x^{(t)})$:
\begin{itemize}
\item Draw $\tilde x^{(t-1)}$ by running $L$ Langevin sampling steps targeting the conditional density $p_\theta (z^{(t-1)}|x^{(t)} )$, with the initial value set to $x^{(t)}$.
\item Draw $\tilde x^{(t)}$ from $\N( \sqrt{\alpha_t} \tilde x^{(t-1)}, (1-\alpha_t)I)$.
\end{itemize}
The two operations constitute one step of Metropolis within Gibbs sampling (or MCMC within Gibbs sampling):
sample $\tilde x^{(t-1)}$ given $x^{(t)}$ using Langevin sampling, and then sample $\tilde x^{(t)}$ given $\tilde x^{(t-1)}$.
In fact, drawing $\tilde x^{(t)}$ given $\tilde x^{(t-1)}$ can be skipped,
because the expectation term in (\ref{eq:bivariate-grad}) involves only $z^{(t-1)}$, but not $z^{(t)}$.
From the preceding discussion, the approximate gradient (\ref{eq:bivariate-grad-approx})
is the same as the approximate gradient (\ref{eq:DRLK-grad-approx}): not only the two expressions are of the same form,
but also $\tilde x^{(t-1)}$ is generated in the same way given $x^{(t)}$.
In this sense, the recovery likelihood approach of \citet{gao2020Diffusion} is equivalent to training the ``bivariate" EBM $p_\theta( x^{(t-1)}, x^{(t)} )$ via CD.

\clearpage

\section{Experiment details and additional results}

\subsection{Gaussian mixture 1D example}\label{sec:append-Gaussian}

We train the EBM with persistent initialization and MALA sampling in the 1D Gaussian mixture example. The energy function is parameterized as $U_\theta(x) = (x/10)^2 + \theta^\T h_\theta(x)$, where the ReLU basis function $h_\theta(x)$ is
\begin{align*}
    h_\theta(x) = ((x-\xi_1)_+, \dots,
   (x-\xi_K)_+ ) ^\T,
\end{align*}
where $\xi_j$'s are the knots and $a_+=a$ if $a\ge 0$ or $0$ if $a<0$.
Equi-spaced knots from $-4$ to $4$ by $.1$ are used, as mentioned in the main text. The prior quadratic term $(x/10)^2$ is needed to guarantee the energy function is well-defined, i.e., the corresponding density function is integrable.  We train the model for 1,200 iterations with a starting learning rate $\gamma=0.2$. Learning rate decays by a factor of 0.2 at the 600, 800, and 1,000 iteration milestones to ensure convergence.
The Langevin step size $\sigma$ is set to $0.1$, and the number of Langevin steps $L$ is set to 10 per training iteration. The replay buffer size is set to 1,000 (same as the training dataset).

\begin{figure}[!h]
\begin{center}
\centerline{\includegraphics[width=\columnwidth]{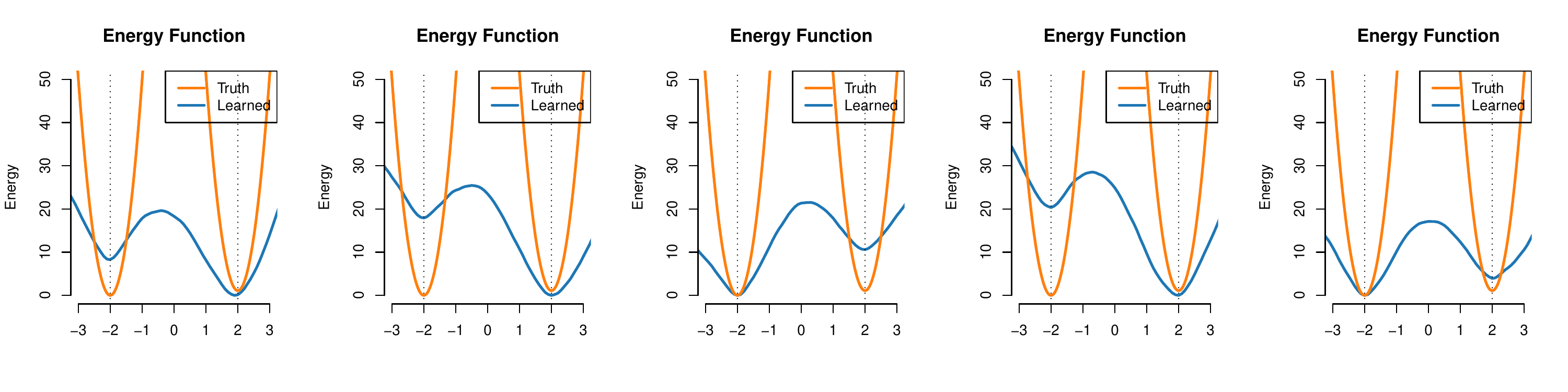}}  \vskip -.2in
\caption{Learned energy functions from five independent runs by persistently training EBMs in 1D example}
\label{fig:Gaussian1D-additional}
\end{center}
\vskip -0.2in
\end{figure}

We plot the estimated energy functions of five more independent runs in Figure \ref{fig:Gaussian1D-additional}. The estimated energies near $-2$ and $2$
exhibit varying directions of relative magnitudes in different training runs. This shows the difficulty in learning a global energy function using persistent training of EBM in the presence of separated modes by low-density barriers.

\subsection{Four-ring example}\label{sec:append-four-ring}

\textbf{Training data.}
For a 2D vector $x=(x_1,x_2)$, suppose that the radius $r = \sqrt{x_1^2+x_2^2}$ of a ring follows a one-sided truncated normal distribution to $(0,\infty)$, and the angle $\theta=\arctan(x_1/x_2)$ follows a uniform distribution over $[0,2\pi]$. The means of the radii of the four rings are $1, 2, 3, 4$, and the standard deviations are 0.01. We draw a total of 50,000 points from the four rings with equal probability as the training set.

\textbf{Network.}
We choose a four-layer MLP, which has 128 equal hidden neurons in each of the middle layers. EBMs trained with different initialization use ReLU as the activation function, while DA-EBM and DRL use the Softplus activation function.  We use sinusoidal time embedding and set $T=6$ for DA-EBM and DRL. Both methods use the same cumulative-sum diffusion scheduling described in Section \ref{sec:app_images_impl} with $\delta_1 =0.01$ and $\delta_T=0.3$.

\textbf{Training and sampling.}
We train all methods with a multi-step learning rate decay schedule. DRL are trained with 500 epochs, and all other methods are trained with 200 epochs. The decay milestones are at 7/10, 8/10, and 9/10 of the total training epoch, and the decay factor is 0.1.
The initial learning rate is set to $5\times 10^{-4}$. The DRL training is strictly based on Algorithms 1 and 2 in \citet{gao2020Diffusion}, without the additional variations in the image experiments (see Section \ref{sec:app_images_impl}),

All EBM methods except our proposed DA-EBM use Langevin sampling without acceptance-rejection by implementation conventions in the literature. For EBM training, we consider two different Langevin step sizes, $\sigma=0.005$ and $\sigma =0.01$, and pick up the best result for illustration. For the DRL method,
we set the step size $\sigma_t = b \sqrt{1-\alpha_{t+1}^2}$ for $t=0, \dots, T-1$ with two possible choices $b=0.02$ and $b=0.01$, and present the best result. The step size of DA-EBM is dynamically adjusted depending on the average acceptance rate for each time label over each epoch. The adjustment method is described in Section \ref{sec:app_images_impl}. The initial values are set to  $\sigma_t = b \sqrt{1-\alpha_{t}^2}$ for $t=1, \dots, T$ and $\sigma_0 = b\sqrt{1-\alpha_{1}^2}$ with $b=0.1$.
During training per iteration, the number of Langevin steps $L$ is set to $L=200$ for hybrid initialization and $L=4000$ for noise initialization due to training instability. For all other methods, $L$ is set to $L=40$.

\begin{table}[!h]
      \caption{Additional results in the four-ring example}
      \label{tb:toy_example_ctd}
    \vskip 0.15in
     \begin{center}
     \begin{tabular}{ c | c    }
     \toprule[1pt]

     {\small\makecell{
     Diffusion \\
     Data Info. } }
       &
       \begin{minipage}{.85\textwidth}
         \includegraphics[width=\linewidth, height=25mm]{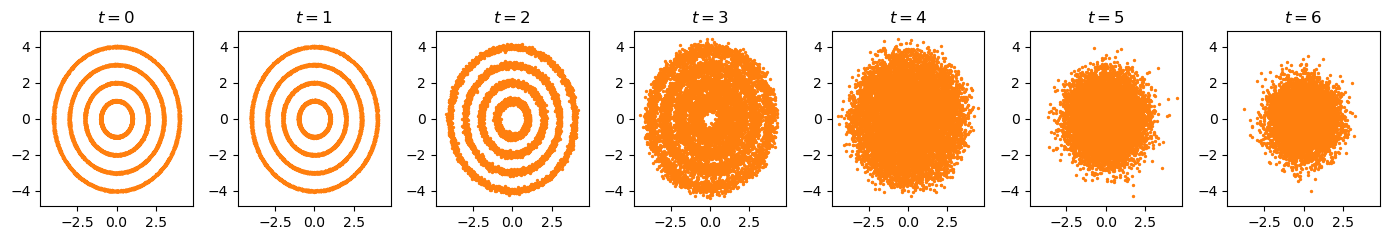}
        \end{minipage}\\
      \cmidrule[1pt](r){1-1}\cmidrule[1pt](l){2-2}
      Method &  Performance \\
      \cmidrule[1pt](r){1-1}\cmidrule[1pt](l){2-2}

     {\small\makecell{
     DRL} } &
        \begin{minipage}{.85\textwidth}
         \includegraphics[width=\linewidth, height=25mm]{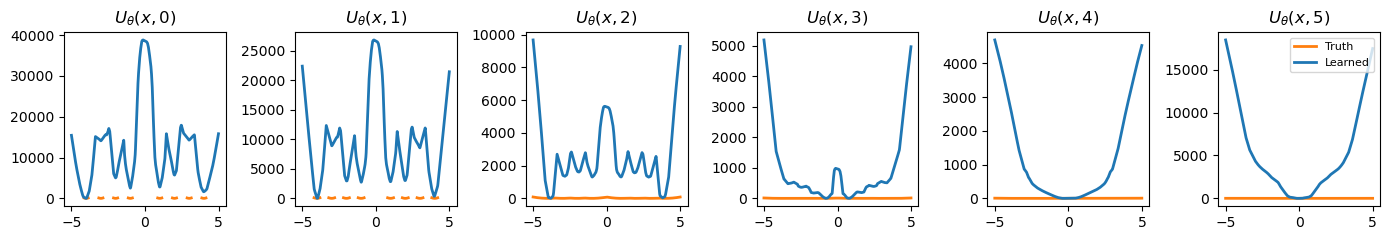}
        \end{minipage}\\ \hline

     {\small\makecell{
     DA-EBM} } &
        \begin{minipage}{.85\textwidth}
         \includegraphics[width=\linewidth, height=25mm]{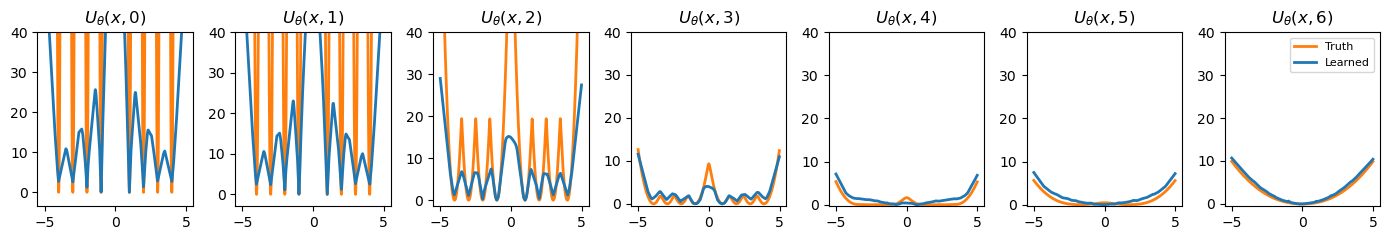}
        \end{minipage}\\
       \bottomrule[1pt]
      \end{tabular}
      \end{center}
  \end{table}

\textbf{Long-run sampling and post-training sampling.}
After training, long-run samples are obtained after 100k Langevin steps, in parallel, starting from an independent draw of 10,000 points from the four-ring distribution, while post-training samples are obtained starting from 10,000 standard Gaussian noises.
For post-training sampling, EBMs trained with noise and hybrid initialization use $L$ and $20L$ Langevian steps. EBMs trained with data initialization and persistent initialization use 10,000 Langevian steps. DRL uses $6L$ Langevian steps in the sequential conditional sampling for post-training sampling. Our DA-EBM runs 10,000 parallel chains for a total of 250 MGMS sampling transitions (Algorithm \ref{alg:joint-sampling}), and time 0 samples are identified as post-training samples.

\textbf{Additional results.}
We demonstrate the learned energy functions for all different time steps in Table \ref{tb:toy_example_ctd}.
For DRL, the energy function at time $6$, $U_\theta(x,6)$, is assumed to be that of standard Gaussian.
As can be seen, our DA-EBM learns the energy functions $U_\theta(x,t)$ that reasonably match with the true negative log densities (up to an additive constant) of diffusion data across all time steps $t=0,\dots, 6$, while DRL fails to do that.

\subsection{Image experiments}\label{sec:app_images_impl}

\vspace{-.1in}

\begin{table}[!h] 
\caption{Model architecture for the experiments on the image data}
\label{tb:network}
\vskip 0.15in
\begin{subtable}[h]{0.33\textwidth}
\subcaption{Network}
\begin{center}
\begin{small}
\begin{sc}
\begin{tabular}{ c  }\toprule
3x3 Conv2D, 128\\\hline
\makecell{1 ResBlock, 128\\
Downsample 2$\times2$}\\ \hline
\makecell{1 ResBlock, 256\\
Downsample 2$\times2$}\\ \hline
\makecell{1 ResBlock, 256\\
Downsample 2$\times2$}\\ \hline
1 ResBlock, 256\\\hline
\makecell{S{\normalfont i}LU, global sum\\\hline
+ Dense(S{\normalfont i}LU({\normalfont temb}))\\\hline
SUM}
\\ \bottomrule
\end{tabular}%
\end{sc}
\end{small}
\end{center}
\end{subtable}
\begin{subtable}[h]{0.33\textwidth}
\subcaption{ResBlock}
\begin{center}
\begin{small}
\begin{sc}
\begin{tabular}{ c  }\toprule
S{\normalfont i}LU, 3$\times$3 Conv2D\\\hline
+ Dense(S{\normalfont i}LU({\normalfont temb}))\\\hline
SiLU, 3$\times$3 Conv2D\\\hline
+  Conv2D(input)
\\ \bottomrule
\end{tabular}%
\end{sc}
\end{small}
\end{center}
\end{subtable}
\begin{subtable}[h]{0.33\textwidth}
\subcaption{Time embedding (temb)}
\begin{center}
\begin{small}
\begin{sc}
\begin{tabular}{ c  }\toprule
Sinusoidal Embedding\\\hline
Dense, S{\normalfont i}LU\\ \hline
Dense
\\ \bottomrule
\end{tabular}%
\end{sc}
\end{small}
\end{center}
\end{subtable}
\vskip -0.1in
\end{table}

\clearpage

\begin{table}[!h]
\caption{Main differences in training implementations of different methods}
\label{tb:hyparam}
\vskip 0.15in
\begin{minipage}{\textwidth}
\begin{center}
\begin{small}
\begin{sc}
\begin{tabular}{ c | c c c c c c }\toprule
 &  $\beta_1$ in Adam &  lr $\gamma$ & Normalization & Image Preprossessing  & $T$ & Number of Param. \\\midrule
EBM Pers.  & 0.0 & $2\times 10^{-5}$ & - & Gaussian (std=0.03) & - & 4.7M\\
EBM Hybrid & 0.0 & $2\times 10^{-5}$&- & Gaussian (std=0.03) & -& 4.7M \\
EBM Noise & 0.0 & $5\times 10^{-6}$&- & Gaussian (std=0.03) & -& 4.7M \\
DRL        & 0.9 & $2\times 10^{-4}$& Spectral  & None & 50 & 4.7M \\
DA-EBM     &  0.0& $1\times 10^{-5}$ & - & Uniform & 50& 4.7M \\
Glow       &  0.9 \footnote{Adamax is used instead of Adam.} & $5\times 10^{-4}$ & Actnorm &  Gaussian (std=0.01) &-  &29.5M  \\
DDPM       & 0.0  & $2\times 10^{-4}$ &  Group  & Uniform  & $[0,999]$  & 20.1M
\\ \bottomrule
\end{tabular}%
\end{sc}
\end{small}
\end{center}
\end{minipage}

\vskip -0.1in
\end{table}

\textbf{Model architecture.}
We adopt a similar network architecture as in \citet{gao2020Diffusion}, which is a variation of the WideResNet \cite{zagoruyko2016wideresnet} and agrees with the bottom-up part of the U-Net architecture used in \citet{ho2020ddpm}. We use uses the Sigmoid Linear Unit (SiLU) as the activation function. Time label $t$ is encoded by Transformer sinusoidal positional embedding \cite{vaswani2017attention}. We list the network structure in Table \ref{tb:network}.

\textbf{Training basics.}
We train all EBM methods with Adam optimizer \cite{kingma2015adam} for 120k iterations, with a mini-batch size of 200. The first 6k iterations are set as the warm-up stage, during which the learning rate linearly increases from 0 to the desired learning rate. The last 48k iterations are set as the annealing stage, during which the learning rate linearly decreases to a small factor of its original level. The linear decay factor is set to $10^{-5}$. We currently use Kaiming initialization \cite{he2015init}. Image is rescaled to range $[-1, 1]$, and additional Gaussian noise or uniform dequantization is added to stabilize training for different methods, as detailed in Table \ref{tb:hyparam}.

We find out that, empirically, persistent training does not work with normalization (e.g., batch/spectral normalization). 
Currently, we remove all normalizations in the network for EBM training methods, except that we keep the spectral normalization for the DRL method. Otherwise, the training fails to learn meaningful models or diverges easily.

\textbf{DRL, Glow, cDDPM.}
DRL training is based on the released code\footnote{\url{https://github.com/ruiqigao/recovery_likelihood}} of the $T6$ setting in \citet{gao2020Diffusion}. Note that the code involves some additional variations from the method described in Section 3.5 of the paper. First, in the code, the energy function of the second last diffusion time $U_\theta(x, T-1)$ is trained in the same way as using ML with noise initialization instead of using recovery likelihood for $x^{(T-1)}$ given $x^{(T)}$.
Second, the Langevin update for drawing from $p_\theta(x^{(t-1)}|x^{(t)})$ is modified as follows with the energy $U_\theta(x,t-1)$ scaled by $\sigma_{t-1}^2$:
\begin{align}
\tilde{x}^{(t-1)}_l = \tilde{x}^{(t-1)}_{l-1} - \frac{\sigma_{t-1}^2}{2}\left\{ \frac{\nabla_x U_\theta(\tilde x_{l-1}^{(t-1)}, t-1) }{ \sigma_{t-1}^2} + \frac{x^{(t)} - \sqrt{\alpha_t}\tilde{x}^{(t-1)}_{l-1} }{1-\alpha_t} \right\} + \sigma_{t-1} \varepsilon , \label{eq:DRL-langevin}
\end{align}
for $l=1,\ldots,L$, with the initial value $\tilde x_0^{(t-1)} = x^{(t)}$ and the Langevin step size $\sigma_t = bc_t \sqrt{1-\alpha_{t+1}^2}$ for an additional factor $c_t$.
This is not proper Langevin dynamics for the conditional EBM $ p_\theta (x^{t-1)} | x^{(t)})$ in Eq.~(16) of  \citet{gao2020Diffusion}.
Finally, a time-dependent weight factor is added to the gradient in updating the parameter. The effects of these additional variations are unclear, but we keep them in our training implementation; otherwise training tends to break down. The parameter $b$ is set to 0.02 and other parameters remain as the default in the DRL code.
For DRL, we keep the improper Langevin update as (\ref{eq:DRL-langevin}) in sequential conditional sampling for post-training image generation,
whereas we use the (proper) Langevin update for the learned energy $U_\theta(x, 0)$ without rescaling to investigate long-run sampling.

The Glow model used in the OOD experiments is trained using the GitHub repository, Glow-PyTorch\footnote{\url{https://github.com/y0ast/Glow-PyTorch}}. We set the mini-batch size to 100 and use the additive coupling layer. Other parameters remain as the default training options in the repository. We also need to add Gaussian noise to the image for preprocessing. Otherwise, log-likelihoods evaluated on the test set get infinity or NaN values. The cDDPM (continuous-time DDPM) model is trained mainly based on the released code of \citet{song2021sde}, score\_sde\_pytorch \footnote{\url{https://github.com/yang-song/score_sde_pytorch}}. We train the cDDPM for 600k iterations with the continuous-time objective function (7) in \citet{song2021sde}. The continuous time step avoids ad-hoc interpolation when evaluating the model density by solving the probability flow SDE. We remove the attention layer in the network structure and choose the network such that the bottom-up part agrees with our network structure in Table \ref{tb:network}.

The main differences in training implementations for different methods are summarized in Table \ref{tb:hyparam}.
Normalized density estimates are directly available from Glow, and
can be obtained by solving a probability flow ODE for cDDPM.

\textbf{Langevin sampling.}
The replay buffer size is set to 50,000.
The number of Langevin steps  $L$ is set to $L=80$ for the noise-initialized EBM and $L=40$ for all other methods. EBMs trained with different initializations all use a Langevin step size 0.005. For our DA-EBM, the step sizes for different time labels are initialized to be 0.01. We turn on acceptance-rejection in the MGMS algorithm after the warm-up stage and adjust step sizes $\sigma_t$ dynamically based on acceptance rates as follows.

The average acceptance rate for each time label is calculated every 100 iterations during training, and the corresponding step size is then adjusted.
Similar to the step size tuning in Appendix V.4 in \citet{song2021hams}, we adjust the step size such that the acceptance rate of each time label is between 0.6 and 0.8 during the training. When the acceptance rate is smaller than 0.6, we decrease $\sigma_t$ by a factor of $1/(1+2\tau)$, and when the acceptance rate is larger than 0.8, we increase $\sigma_t$ by a factor of $(1+\tau)$, where $\tau$ is an adjustment value taken to be $\tau = 0.1$ in all our experiments. The increase and decrease factors are designed to be asymmetric. Otherwise, the step sizes fall in a fixed set of values determined by the initial value.

\textbf{Long-run sampling and post-training sampling.}
We report 100k or 500k long-run samples, obtained after running 100k or 500k Langevin sampling steps, starting from real images,
with the learned energy function. For DRL and DA-EBM, this is the learned energy function $U_{\hat\theta}(x,0)$ at time 0.
Post-training samples are obtained starting from standard Gaussian noises. EBMs trained with noise and hybrid initializations use $L$ and $20L$ Langevian steps for post-training image generation, while EBM trained with persistent initialization uses 100k Langevian steps. DRL uses $50L$ Langevian steps in sequential conditional sampling. Our post-training image generation procedure is described below.

For post-training sampling, the Langevin step size is fixed to be the final step size during training. We start Markov chains from standard Gaussian noises and repeatedly run the MGMS algorithm (Algorithm \ref{alg:joint-sampling}) until MCMC samples become realistic images. In this work, we regard a sampled configuration $x$ as a realistic image when sampled configurations have been  classified as time label $t= 0$ for 50 times in the repeated MGMS runs (see Algorithm~\ref{alg:post-sampling}).

\textbf{Diffusion scheduling.}
While DRL and cDDPM use original scheduling in their released codes, DA-EBM uses a cumulative-sum diffusion scheduling as follows.
We set $\alpha_1,\dots, \alpha_T$ such that the standard deviation  $\sqrt{1-\alpha_t}$ in the diffusion process is increasing with respect to $t$ as a cumulative sum.  Specifically, we set $T=50$ and set $\delta_i$ to increase from $\delta_1=0.0002$ to $\delta_T = 0.02$ with equal spaces. Then we determine $\alpha_t$ according to $\sqrt{1-\alpha_t} = \sum_{i=1}^t\delta_i$. The diffusion scheduling is not particularly optimized and can possibly be improved for better training performance.

\textbf{Additional sampling results.}
The results of (500k) long-run sampling and image generation on MNIST are presented in Table \ref{tb:append_mnist_table500k} for noise initialized EBM
as well as for other methods (replicated from Table \ref{tb:mnist_imgs} for completeness).
The results of (500k) long-run sampling and image generation on FashionMNIST are presented in Table \ref{tb:fmnist_imgs_table500k}.
We also show additional long-run samples obtained from 100k MCMC steps in Table \ref{tb:app_longrun100k}. The results are similar to 500k long-run samples, but may be less obvious for visual inspection. Additional post-training samples of DA-EBM are presented in Table \ref{tb:app_addtional_post}.

\begin{figure}[ht]
\vskip 0.2in
\begin{center}
\begin{subfigure}[b]{\textwidth}
\includegraphics[width=\textwidth]{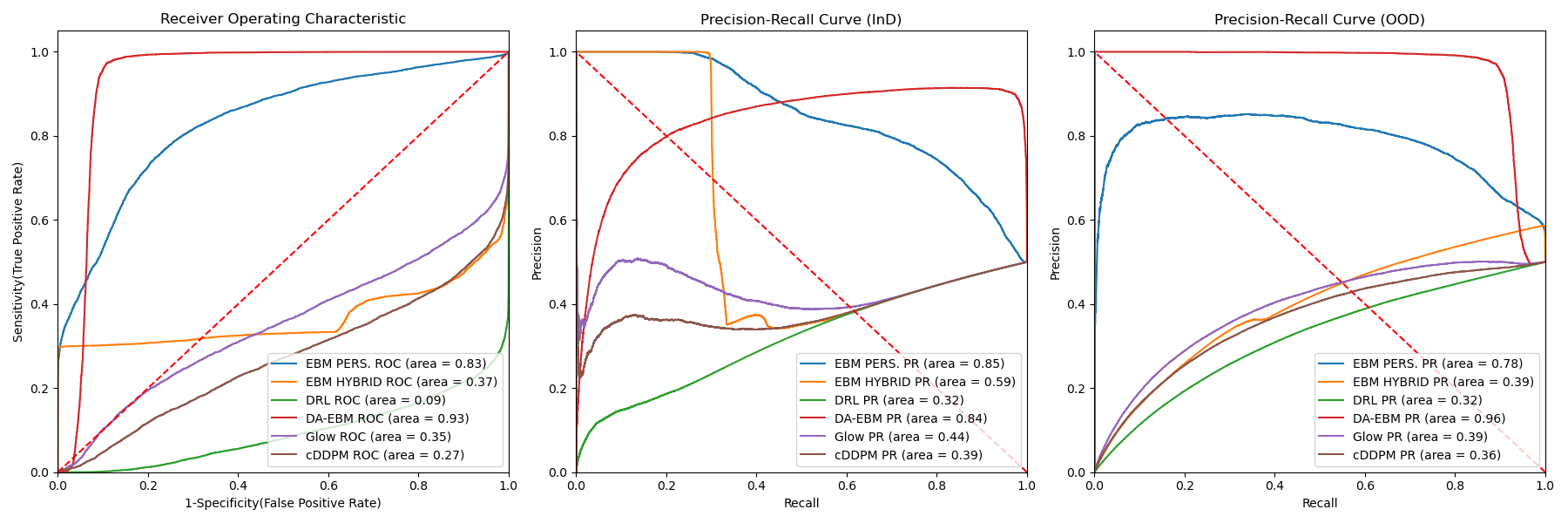}
\caption{Fashion-MNIST (InD) vs MNIST(OOD)}
\end{subfigure}
\begin{subfigure}[b]{\textwidth}
\includegraphics[width=\textwidth]{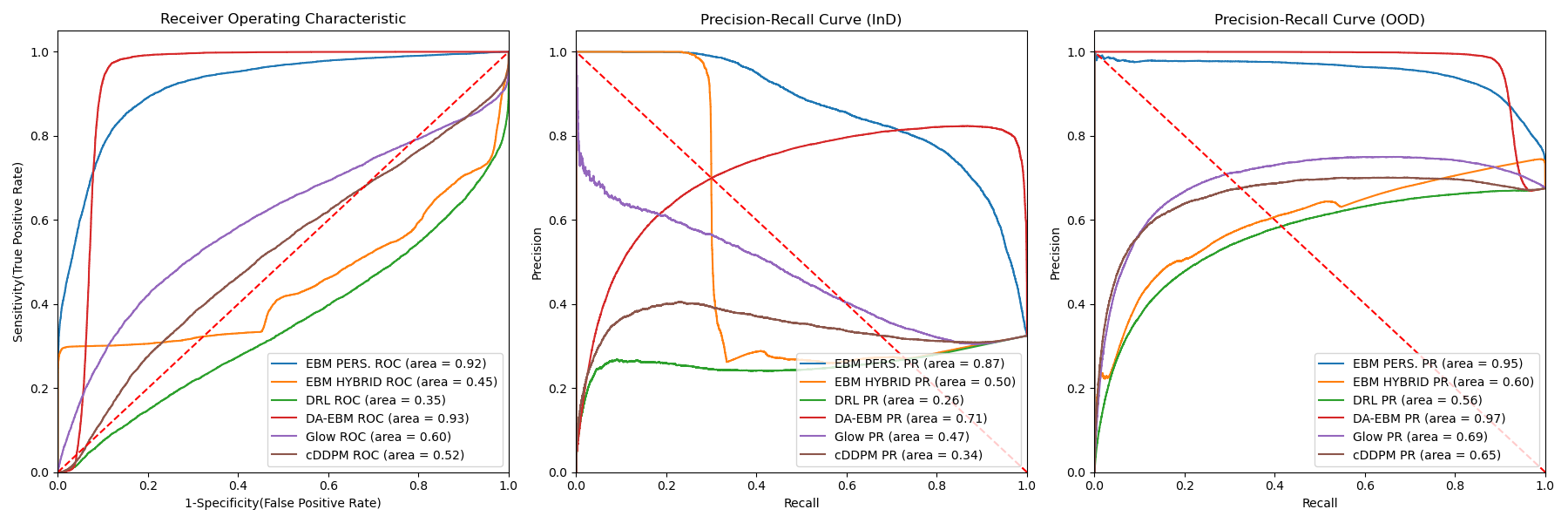}
\caption{Fashion-MNIST (InD) vs EMNIST(OOD)}
\end{subfigure}
\begin{subfigure}[b]{\textwidth}
\includegraphics[width=\textwidth]{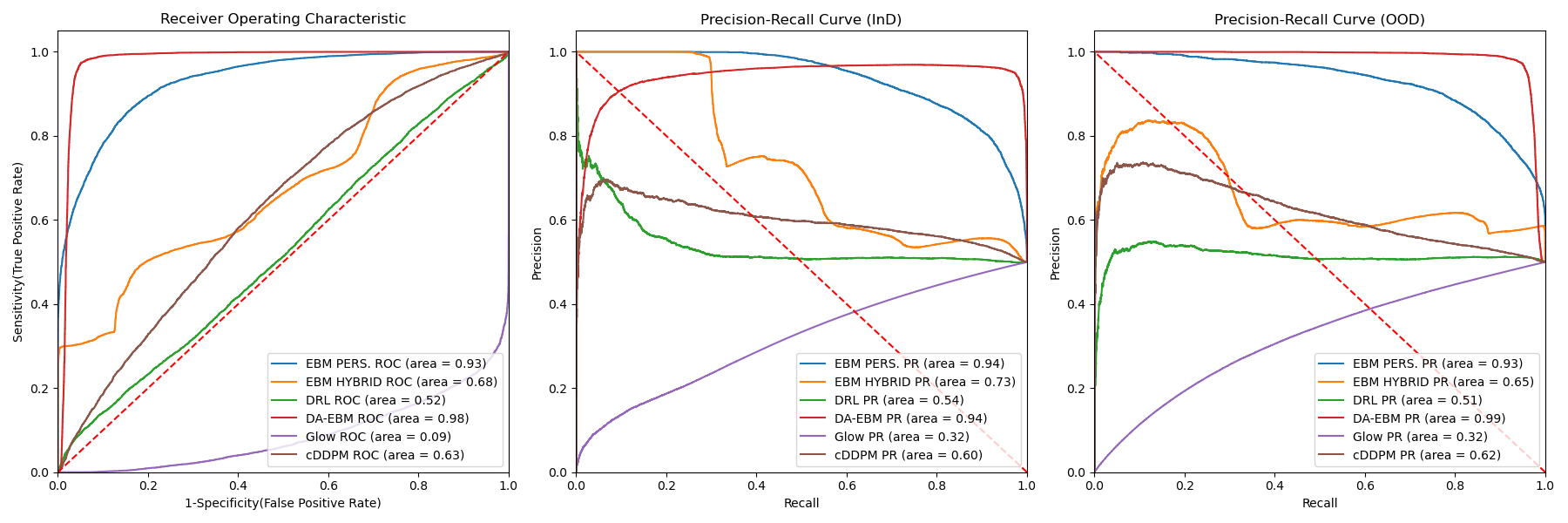}
\caption{Fashion-MNIST (InD) vs KMNIST(OOD)}
\end{subfigure}
\caption{ROC curves, PR curves}
\label{fig:roc_pr_curves}
\end{center}
\vskip -0.2in
\end{figure}

\textbf{Additional OOD results.}
We plot the ROC curves and PR curves in Figure \ref{fig:roc_pr_curves} and summarize the AUC-PR results in Table \ref{tb:ood_aucpr_results}. A comparison of the computational cost of likelihood evaluation for DA-EBM and cDDPM is shown in Table \ref{tb:ood_computation}. The computation time for cDDPM is huge compared to the almost negligible time for DA-EBM, which shows the advantage of learning EBMs for OOD detection.

\begin{table*}[!ht]
\caption{Results of (500k) long-run sampling and image generation on FashionMNIST. A replica of Table \ref{tb:mnist_imgs} with additional results of noise initialized EBM.}
\label{tb:append_mnist_table500k}
\vskip 0.15in
\begin{center}
\begin{small}
\begin{sc}
     \begin{tabular}{ c | c | c  }
     \toprule[1pt]
     & \multicolumn{2}{c}{mnist} 
        \\\hline
      \makecell{Train. Data \\
        \scriptsize (Long-run starting points)
      }
      &      
      \multicolumn{2}{c}{    \begin{minipage}{.25\columnwidth}
         \includegraphics[width=\linewidth, height=21mm]{final_paper_plots/mnist-rep-imgs.png}
        \end{minipage} }
        \\
      \cmidrule[1pt](r){1-1}\cmidrule[1pt](lr){2-2} \cmidrule[1pt](l){3-3}
      Method &  Long-run & Post  \\
      \cmidrule[1pt](r){1-1}\cmidrule[1pt](lr){2-2} \cmidrule[1pt](l){3-3} 
      {\small  \makecell{ EBM 
    Pers. 
    }}   &
        \begin{minipage}{.25\columnwidth}
         \includegraphics[width=\linewidth, height=21mm]{final_paper_plots/mnist/odebm_ps_longrun500k.png}
        \end{minipage} &
        \begin{minipage}{.25\columnwidth}
         \includegraphics[width=\linewidth, height=21mm]{final_paper_plots/mnist/odebm_ps_post.png}
        \end{minipage} 
        \\\cmidrule[.2pt]{1-3}
      {\small  \makecell{  EBM 
       Noise }}   &
        \begin{minipage}{.25\columnwidth}
         \includegraphics[width=\linewidth, height=21mm]{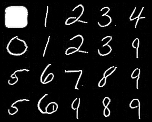}
        \end{minipage} &
        \begin{minipage}{.25\columnwidth}
         \includegraphics[width=\linewidth, height=21mm]{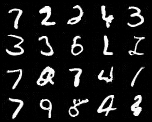}
        \end{minipage} 
          \\\cmidrule[.2pt]{1-3}
      {\small  \makecell{EBM 
    Hybrid \\
    (5\% noise) }}   &
        \begin{minipage}{.25\columnwidth}
         \includegraphics[width=\linewidth, height=21mm]{final_paper_plots/mnist/odebm_restart_longrun500k.png}
        \end{minipage} &
        \begin{minipage}{.25\columnwidth}
         \includegraphics[width=\linewidth, height=21mm]{final_paper_plots/mnist/odebm_restart_post.png}
        \end{minipage} 
        \\\cmidrule[.2pt]{1-3}
      {\small  \makecell{ DRL }}   &
        \begin{minipage}{.25\columnwidth}
         \includegraphics[width=\linewidth, height=21mm]{final_paper_plots/mnist/drlk_longrun500k.png}
        \end{minipage} &
        \begin{minipage}{.25\columnwidth}
         \includegraphics[width=\linewidth, height=21mm]{final_paper_plots/mnist/drlk_post.png}
        \end{minipage} 
        \\\cmidrule[.2pt]{1-3}
      {\small  \makecell{ DA-EBM }}   &
        \begin{minipage}{.25\columnwidth}
         \includegraphics[width=\linewidth, height=21mm]{final_paper_plots/mnist/daebm_longrun500k.png}
        \end{minipage} &
        \begin{minipage}{.25\columnwidth}
         \includegraphics[width=\linewidth, height=21mm]{final_paper_plots/mnist/daebm_post.png}
        \end{minipage}  
        \\\bottomrule[1pt]
  \end{tabular}

\end{sc}
\end{small}
\end{center}
\vskip -0.1in
\end{table*}

\begin{table*}[!ht]
\caption{Results of (500k) long-run sampling and image generation on FashionMNIST}
\label{tb:fmnist_imgs_table500k}
\vskip 0.15in
\begin{center}
\begin{small}
\begin{sc}
     \begin{tabular}{ c | c | c  }
     \toprule[1pt]
     & \multicolumn{2}{c}{FashionMnist} 
        \\\hline
      \makecell{Train. Data \\
        \scriptsize (Long-run starting points)
      }
      &      
      \multicolumn{2}{c}{    \begin{minipage}{.25\columnwidth}
         \includegraphics[width=\linewidth, height=21mm]{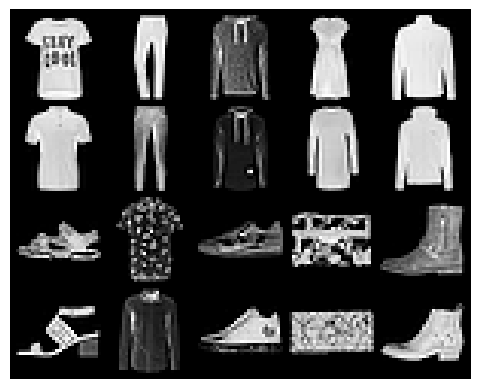}
        \end{minipage} }
        \\
      \cmidrule[1pt](r){1-1}\cmidrule[1pt](lr){2-2} \cmidrule[1pt](l){3-3}
      Method &  Long-run & Post  \\
      \cmidrule[1pt](r){1-1}\cmidrule[1pt](lr){2-2} \cmidrule[1pt](l){3-3} 
      {\small  \makecell{ EBM 
    Pers. 
    }}   &
        \begin{minipage}{.25\columnwidth}
         \includegraphics[width=\linewidth, height=21mm]{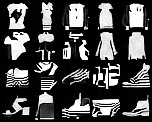}
        \end{minipage} &
        \begin{minipage}{.25\columnwidth}
         \includegraphics[width=\linewidth, height=21mm]{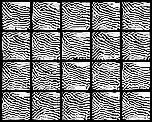}
        \end{minipage} 
        \\\cmidrule[.2pt]{1-3}
      {\small  \makecell{  EBM 
       Noise }}   &
        \begin{minipage}{.25\columnwidth}
         \includegraphics[width=\linewidth, height=21mm]{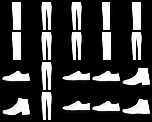}
        \end{minipage} &
        \begin{minipage}{.25\columnwidth}
         \includegraphics[width=\linewidth, height=21mm]{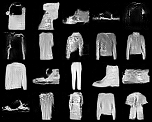}
        \end{minipage} 
          \\\cmidrule[.2pt]{1-3}
      {\small  \makecell{EBM 
    Hybrid \\
    (5\% noise) }}   &
        \begin{minipage}{.25\columnwidth}
         \includegraphics[width=\linewidth, height=21mm]{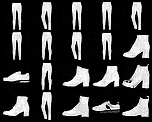}
        \end{minipage} &
        \begin{minipage}{.25\columnwidth}
         \includegraphics[width=\linewidth, height=21mm]{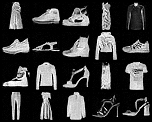}
        \end{minipage} 
        \\\cmidrule[.2pt]{1-3}
      {\small  \makecell{ DRL }}   &
        \begin{minipage}{.25\columnwidth}
         \includegraphics[width=\linewidth, height=21mm]{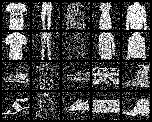}
        \end{minipage} &
        \begin{minipage}{.25\columnwidth}
         \includegraphics[width=\linewidth, height=21mm]{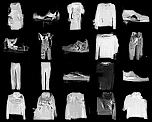}
        \end{minipage} 
        \\\cmidrule[.2pt]{1-3}
      {\small  \makecell{ DA-EBM }}   &
        \begin{minipage}{.25\columnwidth}
         \includegraphics[width=\linewidth, height=21mm]{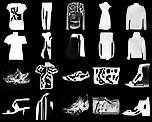}
        \end{minipage} &
        \begin{minipage}{.25\columnwidth}
         \includegraphics[width=\linewidth, height=21mm]{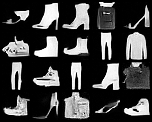}
        \end{minipage} 
        \\\bottomrule[1pt]
  \end{tabular}
\end{sc}
\end{small}
\end{center}
\vskip -0.1in
\end{table*}

\begin{table*}[!ht]
\caption{Results of (100k) long-run sampling on MNIST and FashionMNIST}
\label{tb:app_longrun100k}
\vskip 0.15in
\begin{center}
\begin{small}
\begin{sc}
      \begin{tabular}{ c | c || c }
     \toprule[1pt]
     & \multicolumn{1}{c||}{MNIST} &\multicolumn{1}{c}{Fashion-Mnist} 
        \\\hline
         \makecell{Train. Data \\
        \scriptsize (Long-run starting points)
        }
      &      
      \multicolumn{1}{c||}{    \begin{minipage}{.25\textwidth}
         \includegraphics[width=\linewidth, height=25mm]{final_paper_plots/mnist-rep-imgs.png}
        \end{minipage}  } 
        &
        \multicolumn{1}{c}{   \begin{minipage}{.25\textwidth}
             \includegraphics[width=\linewidth, height=25mm]{final_paper_plots/fmnist-rep-imgs.png}
        \end{minipage} }
        \\
      \cmidrule[1pt](r){1-1}\cmidrule[1pt](lr){2-2} \cmidrule[1pt](l){3-3}
      Method &  Long-run & Long-run \\
      \cmidrule[1pt](r){1-1}\cmidrule[1pt](lr){2-2} \cmidrule[1pt](l){3-3}
      {\small  \makecell{ EBM PERS.
    }}   &
        \begin{minipage}{.25\textwidth}
         \includegraphics[width=\linewidth, height=25mm]{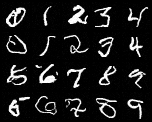}
        \end{minipage} &
        \begin{minipage}{.25\textwidth}
         \includegraphics[width=\linewidth, height=25mm]{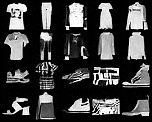}
        \end{minipage} 
        \\\cmidrule[.2pt]{1-3}
      {\small  \makecell{  EBM NOISE}}   &
        \begin{minipage}{.25\textwidth}
         \includegraphics[width=\linewidth, height=25mm]{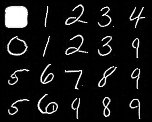}
        \end{minipage} &
       \begin{minipage}{.25\textwidth}
         \includegraphics[width=\linewidth, height=25mm]{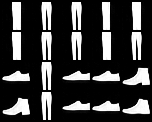}
        \end{minipage} 
        \\\cmidrule[.2pt]{1-3}
      {\small  \makecell{ EBM HYBRID \\
    (5\% noise) }}   &
        \begin{minipage}{.25\textwidth}
         \includegraphics[width=\linewidth, height=25mm]{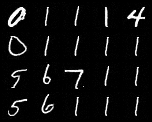}
        \end{minipage} &
       \begin{minipage}{.25\textwidth}
         \includegraphics[width=\linewidth, height=25mm]{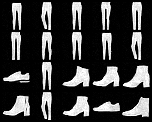}
        \end{minipage} 
        \\\cmidrule[.2pt]{1-3}        
      {\small  \makecell{ DRL }}   &
        \begin{minipage}{.25\textwidth}
         \includegraphics[width=\linewidth, height=25mm]{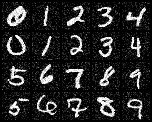}
        \end{minipage} &        
        \begin{minipage}{.25\textwidth}
         \includegraphics[width=\linewidth, height=25mm]{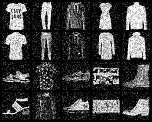}
        \end{minipage} 
        \\\cmidrule[.2pt]{1-3}
      {\small  \makecell{ DA-EBM }}   &
        \begin{minipage}{.25\textwidth}
         \includegraphics[width=\linewidth, height=25mm]{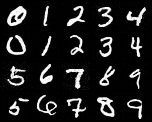}
        \end{minipage} &
        \begin{minipage}{.25\textwidth}
         \includegraphics[width=\linewidth, height=25mm]{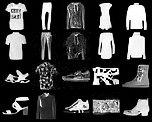}
        \end{minipage} 
        \\\bottomrule[1pt]
  \end{tabular}
\end{sc}
\end{small}
\end{center}
\vskip -0.1in
\end{table*}

\begin{table*}[!ht]
\caption{Additional images generated by DA-EBM in post-training sampling}
\label{tb:app_addtional_post}
\vskip 0.15in
\begin{center}
\begin{small}
\begin{sc}
      \begin{tabular}{ c | c || c }
     \toprule[1pt]
      Method &  MNIST & FashionMnist \\
      \cmidrule[1pt](r){1-1}\cmidrule[1pt](lr){2-2} \cmidrule[1pt](l){3-3}
   
      {\small  \makecell{ DA-EBM }}   &
        \begin{minipage}{.4\textwidth}
         \includegraphics[width=\linewidth, height=60mm]{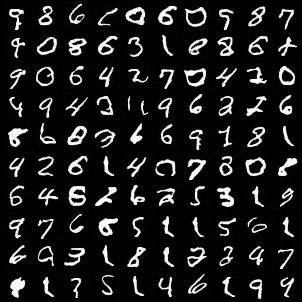}
        \end{minipage} &
        \begin{minipage}{.4\textwidth}
         \includegraphics[width=\linewidth, height=60mm]{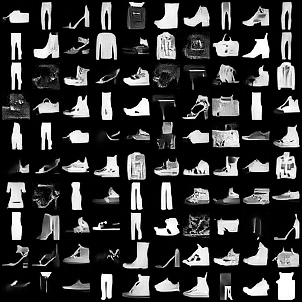}
        \end{minipage} 
        \\\bottomrule[1pt]
  \end{tabular}
\end{sc}
\end{small}
\end{center}
\vskip -0.1in
\end{table*}

\begin{table}[!h]
\caption{Elapsed time for likelihood evaluation on test split of datasets using a single NVIDIA GeForce GTX 1080Ti GPU}
\label{tb:ood_computation}
\vskip 0.15in
\begin{minipage}{\textwidth}
\begin{center}
\begin{small}
\begin{sc}
\begin{tabular}{ c | c c c c c }\toprule
 &  Fashion-MNIST & MNIST & EMNIST & KMNIST  \\\
  \# of samples & 10,000 & 10,000 & 14,800 & 10,000\\\midrule
DA-EBM &  $\approx 2s$ & $\approx 2s$  & $\approx 4s$  & $\approx 2s$   \\
cDDPM & 1h 49m 4s & 2h 6m 49s &  4h 30m 24s  & 2h 11m 17s
\\ \bottomrule
\end{tabular}%
\end{sc}
\end{small}
\end{center}
\end{minipage}

\vskip -0.1in
\end{table}

\begin{table}[t]
\caption{OOD AUC-PR results. InD column indicates that InD examples are
used as the positive class, while OOD column indicates that
OOD examples are used as the positive class.}
\label{tb:ood_aucpr_results}
\vskip 0.15in
\begin{center}
\begin{small}
\begin{sc}
\begin{tabular}{ c | c c c  ||  c c c  }\toprule
& \multicolumn{3}{c||}{InD}& \multicolumn{3}{c}{OOD}\\\midrule
Method &  Mnist  &  EMnist & KMnist &  Mnist  &  EMnist & KMnist \\\midrule
EBM PERS.      & \textbf{0.85} &\textbf{ 0.87 }& \textbf{0.94 }& 0.78 & 0.95 & 0.93 \\
EBM HYBRID     & 0.59 & 0.50 & 0.73 & 0.39 & 0.60 & 0.65\\
DRL             & 0.32 & 0.26 & 0.54 & 0.32 & 0.56 & 0.51\\
DA-EBM         & 0.84 & 0.71 &\textbf{ 0.94 }& \textbf{0.96 }&\textbf{ 0.97} & \textbf{0.99}\\
GLOW           & 0.44 & 0.47 & 0.32 & 0.39 & 0.69 & 0.62\\
cDDPM           & 0.39 & 0.34 & 0.60 & 0.36 & 0.65 & 0.32\\
\bottomrule
\end{tabular}%
\end{sc}
\end{small}
\end{center}
\vskip -0.1in
\end{table}

\section{Limitations}\label{sec:append-limits}

Our post-training sampling is more costly than short-run EBM, hybrid-initialized EBM, and DRL. The average Langevin steps needed for generating 100 images are about 274,162 and 723,844 in Fashion-MNIST and MNIST experiments. 
To scale our method to more complex images and larger models, further investigation is needed.
Nevertheless, we believe this work presents a necessary step towards improving persistent training of EBMs  for complex data like natural images,
to achieve post-sampling generation and proper density estimation simultaneously.

\end{document}